\documentclass[]{meta-template/fairmeta}

\usepackage{amsmath,amsfonts,bm, bbm}

\def\eqref#1{equation~\ref{#1}}

\def\1{\bm{1}}

\def\rvp{{\mathbf{p}}}

\DeclareMathAlphabet{\mathsfit}{\encodingdefault}{\sfdefault}{m}{sl}
\SetMathAlphabet{\mathsfit}{bold}{\encodingdefault}{\sfdefault}{bx}{n}
\newcommand{\tens}[1]{\bm{\mathsfit{#1}}}
\def\tA{{\tens{A}}}

\def\tQ{{\tens{Q}}}

\def\tX{{\tens{X}}}
\def\tY{{\tens{Y}}}
\def\tZ{{\tens{Z}}}

\def\gC{{\mathcal{C}}}
\def\gD{{\mathcal{D}}}

\def\gN{{\mathcal{N}}}

\def\gR{{\mathcal{R}}}
\def\gS{{\mathcal{S}}}

\def\sK{{\mathcal{K}}}

\newcommand{\etens}[1]{\mathsfit{#1}}

\def\etY{{\etens{Y}}}
\def\etZ{{\etens{Z}}}

\def\todo#1{}

\usepackage{xspace}
\usepackage{multirow}
\usepackage{subcaption}
\usepackage{pifont}
\usepackage{graphicx}
\usepackage{booktabs}
\usepackage{pdflscape}
\usepackage[most]{tcolorbox}
\usepackage{enumitem}
\usepackage[normalem]{ulem}

\tcbset{my_style/.style={
  colback=teal!2,
  colframe=teal,
  coltitle=black,
  colbacktitle=white,
  attach boxed title to top left={yshift=-0.1in,xshift=0.15in},
    boxed title style={boxrule=0pt,colframe=white,},
}}

\def\cd{consistency-diversity\xspace}
\def\Cd{Consistency-diversity\xspace}
\def\qd{realism-diversity\xspace}
\def\Qd{Realism-diversity\xspace}
\def\cq{consistency-realism\xspace}
\def\Cq{Consistency-realism\xspace}
\def\cdq{consistency-diversity-realism\xspace}
\def\Cdq{Consistency-diversity-realism\xspace}

\def\coco{MSCOCO2014\xspace}

\def\sdone{LDM$_{1.5}$\xspace}

\def\sdtwo{LDM$_{2.1}$\xspace}
\def\unclip{LDM$_{2.1\text{-UnCLIP}}$\xspace}

\def\sdxl{LDM$_\text{XL}$\xspace}
\def\sdxlt{LDM$_\text{XL-Turbo}$\xspace}
\def\rdm{RDM\xspace}
\def\perco{PerCo\xspace}

\def\clipscore{CLIPScore\xspace}

\def\kscale{g-scale\xspace}
\def\ktopk{top-$m$ filtering\xspace}
\def\kneighbor{$k$-neighbors\xspace}
\def\kcomp{comp. rate\xspace}

\def\igenmodel{image generative model\xspace}
\def\igenmodels{image generative models\xspace}

\def\cigenmodels{conditional image generative models\xspace}
\def\cond{conditioning\xspace}

\def\gen{g_\theta}
\def\enc{f_\phi}

\def\dreamsim{DreamSim\xspace}

\def\dsg{DSG\xspace}

\makeatletter
\DeclareRobustCommand\onedot{\futurelet\@let@token\@onedot}
\def\@onedot{\ifx\@let@token.\else.\null\fi\xspace}

\def\eg{\emph{e.g}\onedot} 
 
\def\cf{\emph{cf}\onedot}

\makeatother

\def\yes{\ding{51}}
\def\no{{\color{lightgray}\ding{55}}}

\newcommand{\topk}{\operatorname{top}}

\tcbset{my_style/.style={
  colupper=metafg,
  colback=metabg,
  colframe=white,
  coltitle=black,
  colbacktitle=white,
  arc=10pt
}}

\title{\Cdq Pareto fronts\\of \cigenmodels}

\author[1]{Pietro Astolfi}
\author[1]{Marlene Careil}
\author[1]{Melissa Hall}
\author[1]{Oscar Mañas}
\author[1]{Matthew Muckley}
\author[1]{Jakob Verbeek}
\author[1, 2, 3, 4, \dagger]{Adriana Romero-Soriano}
\author[1,\dagger]{Michal Drozdzal}

\affiliation[1]{FAIR at Meta - Montreal, Paris, and New York City labs}
\affiliation[2]{McGill University}
\affiliation[3]{Mila, Quebec AI institute}
\affiliation[4]{Canada CIFAR AI chair}

\contribution[\dagger]{Joint last author}

\abstract{Building \emph{world models} that accurately and comprehensively represent the real world is the utmost aspiration for conditional image generative models as it would enable their use as world simulators. For these models to be successful world models, they should not only excel at image quality and prompt-image consistency but also ensure high representation diversity. However, current research in generative models mostly focuses on creative applications that are predominantly concerned with human preferences of image quality and aesthetics. We note that generative models have inference time mechanisms – or \emph{knobs} -- that allow the control of generation consistency, quality, and diversity. In this paper, we use state-of-the-art text-to-image and image\&text-to-image models and their knobs to draw consistency-diversity-realism Pareto fronts that provide a holistic view on consistency-diversity-realism multi-objective. Our experiments suggest that realism and consistency can both be improved simultaneously; however there exists a clear tradeoff between realism/consistency and diversity. By looking at Pareto optimal points, we note that earlier models are better at representation diversity and worse in consistency/realism, and more recent models excel in consistency/realism while decreasing significantly the representation diversity. By computing Pareto fronts on a geodiverse dataset, we find that the first version of latent diffusion models tends to perform better than more recent models in all axes of evaluation, and there exist pronounced consistency-diversity-realism disparities between geographical regions. Overall, our analysis clearly shows that there is \emph{no best model} and the choice of model should be determined by the downstream application. With this analysis, we invite the research community to consider Pareto fronts as an analytical tool to measure progress towards world models.\looseness-1}

\date{\today}
\correspondence{\email{pietroastolfi@meta.com}, \email{mdrozdzal@meta.com}}

\begin{document}

\maketitle

\section{Introduction}
\label{sec:intro}

Progress in foundational vision-based machine learning models has heavily relied on large-scale Internet-crawled datasets of real images~\citep{schuhmann2022laion}. Yet, with the acceleration of research on generative models and the unprecedented photorealistic quality achieved by recent text-to-image generative models~\citep{podell2023sdxl, esser2024scaling, ramesh2022hierarchical, saharia2022photorealistic}, researchers have started exploring their potential as \emph{world models} that generate images to train downstream representation learning models~\citep{astolfi2023instance, hemmat2023feedback, tian2024stablerep}. 

World models aim to represent the real world as accurately and comprehensively as possible. Therefore, visual world models should not only be able to yield \emph{high quality} image generations, but also generate content that is representative of the \emph{diversity} of the world, while ensuring \emph{prompt consistency}. However, state-of-the-art \cigenmodels have mostly been optimized for human preference, and thus, a single high-quality and consistent sample fulfills the current optimization criteria. This vastly disregards representation diversity~\citep{hall2024dig,sehwag2022generating, zameshina2023diverse,corso2024particle,hemmat2023feedback,sadat2024cads}, and questions the potential of state-of-the-art \cigenmodels to operate as effective world models. %
Optimizing for human preferences only partially fulfills the multi-objective optimization required to leverage conditional generative models as world models.\looseness-1

At the same time, state-of-the-art \cigenmodels have built-in inference time mechanisms, hereinafter referred to as \emph{knobs}, to control the realism (also referred to as quality or fidelity), consistency, and diversity dimensions of the generation process. For example, it has been shown that the guidance scale in classifier free guidance of diffusion models~\citep{ho2021classifierfree}, trades image fidelity for diversity~\citep{saharia2022photorealistic, corso2024particle}. Similarly, post-hoc filtering~\citep{karthik2023if} is used to improve consistency. Although recent works have carried out extensive evaluations of image generative models~\citep{ku2024imagenhub,lee2024holistic}, these evaluations have been primarily designed from the perspective of creative applications. 
To the best of our knowledge, a comprehensive and systematic analysis of the effect of the knobs controlling the different performance dimensions of \cigenmodels has not yet been carried out.\looseness-1

In this paper, we benchmark \cigenmodels in terms of the world models' multi-objective. %
In particular, we perform an optimization over both knobs and state-of-the-art models with the goal of capturing the \cd, \qd, and \cq Pareto fronts that are currently reachable.
In our analysis, we include both text-to-image (T2I) models and image\&text-to-image (I-T2I) models. For T2I, we consider several version of latent diffusion models (LDM), namely \sdone and \sdtwo~\citep{rombach2022high}, 
 as well as \sdxl~\citep{podell2023sdxl}, whereas for I-T2I, we consider a retrieval-augmented diffusion model (\rdm)~\citep{blattmann2022semi} and \unclip~\citep{ramesh2022hierarchical}, in addition to a neural image compression model, \perco~\citep{careil2024towards}. 
We perform our T2I and I-T2I models analysis using the ubiquitous MSCOCO~\citep{lin2014microsoft} validation dataset and we extend our evaluation of T2I models to the GeoDE dataset~\citep{ramaswamy2024geode}, composed of images from 6 world regions, to characterize the progress of these models from a geographic representation perspective. To quantify the multi-objective, we use inter-sample similarity and recall~\citep{kynkaanniemi2019improved} to measure representation diversity; image reconstruction quality and precision~\citep{kynkaanniemi2019improved} to quantify realism; and the Davidsonian scene graph score~\citep{cho2024davidsonian}) to assess prompt-generation consistency.\looseness-1

By drawing the Pareto fronts, we observe that progress in \cigenmodels has been driven by improvements in image realism and/or prompt-generation consistency, and that these improvements result in models sacrificing representation diversity. In the T2I setup, our analysis suggests that more recent models should be used when the downstream task requires samples with high realism
-- \sdxlt -- and consistency -- \sdxl --. However, older models -- \sdone and \sdtwo -- are
preferable for tasks that require good representation diversity. For I-T2I models, we observe that compression models --  \eg, \perco -- should be prioritized when working on downstream applications that
require high realism and consistency. However, when the downstream application requires high
representation diversity, \rdm and \unclip are preferable. Interestingly, on GeoDE we observe that the oldest model, \sdone, outperforms the most recent ones, and consistently appears
in the Pareto fronts of all regions considered. Moreover, the advances in T2I models reduce region-wise disparities in terms of consistency and increase the
disparities in terms of image realism, while sacrificing diversity across all regions. Finally, by looking at the knobs, we observe that guidance and post-hoc filtering have the highest effect on the \cd and \qd tradeoffs, increasing both realism and consistency at the expense of representation diversity. 
We believe that the proposed evaluation framework and the findings that arise from it will enable faster progress towards enabling the use of \cigenmodels as world models, and we hope it will encourage the research community to work on models that present softer \cdq tradeoffs.\looseness-1

\section{Methodology of the analysis}
\label{sec:methodology}

In this section, we introduce the notation adopted throughout the rest of the paper, present the metrics we use to evaluate \cigenmodels, and describe existing knobs that control the \cdq multi-objective.\looseness-1

\noindent\textbf{Notation.} Let us consider the following conditional image generation framework. 
An \igenmodel, $\gen$, parameterized by a set of learnable parameters, $\theta$, generates an image, $\etY$, from a noise sample $\etZ\sim\gN(0,\mathbf{I})$ and a \cond prompt encoded by a vector, $\rvp \in \mathbbm{R}^d$: $\etY = \gen(\etZ, \rvp)$. In state-of-the-art \cigenmodels, $\rvp$ encodes either text, an image, or a combination of both. In practice, images are generated in batches of $N$ elements, $\tY \in \mathbbm{R}^{N\times H\times W\times 3}$, conditioned on the \emph{same} vector $\rvp$ and a tensor $\tZ \in \mathbbm{R}^{N\times H\times W\times 3}$ representing $N$ random noise samples:
\begin{equation}
    \tY = \gen(\tZ, \rvp).
    \label{eq:cond_gen}
\end{equation}

\subsection{Evaluating conditional image generation}
\label{sec:metrics}

We evaluate conditional image generation in terms of prompt-sample consistency, sample diversity and realism (also referred to as quality or fidelity in the literature). We consider two complementary ways of 
quantifying the performance of \cigenmodels: conditional and marginal. On the one hand, \emph{conditional metrics} are prompt-specific scores computed on the set of image generations resulting from a prompt. An overall score may be obtained by averaging out all prompt-specific scores. On the other hand, \emph{marginal metrics} are overall scores computed on the generations resulting from \emph{all} the prompts directly. In practice, marginal metrics compare a set of generated images to a reference dataset while ignoring the prompts used to obtain the sets. In the reminder of this subsection, we define consistency -- that is always conditional--, conditional and marginal diversity, as well as conditional and marginal realism.\looseness-1

\noindent\textbf{Consistency, $\gC$.} Prompt-generation consistency is measured either with distance or similarity-based scores -- \eg, CLIPScore~\citep{hessel2021clipscore}, LPIPS score~\citep{zhang2018unreasonable} and DreamSim score~\citep{fu2023dreamsim} -- or with visual question answering (VQA) approahces -- \eg,  TIFA~\citep{hu2023tifa} and DSG~\citep{cho2024davidsonian} metrics--.  In our analysis, we opt to use VQA approaches as they are reported to be more calibrated and interpretable than the distance and similarity-based scores~\citep{cho2024davidsonian}. Concretely, we measure the prompt-generation consistency with DSG. DSG relies on questions $\tQ$ generated from the prompt $\rvp$ and their corresponding answers $\tA$. Per-prompt consistency, $\gC^{p}$, is defined as follows:  %
\begin{equation}
    \gC^{p} = \frac{1}{N}\sum_{j=1}^{N} \frac{1}{Q_j}\sum_{i=1}^{Q_j}{\mathbbm{1}\Big(\mathrm{VQA}(\tY_{j}, \tQ_{i}}), \tA_{i}\Big),
    \label{eq:consistency}
\end{equation} 
where $N$ represents the number of images generated per conditioning prompt, $Q_j$ represents the number of question per j-th image, %
and $\mathbbm{1}$ represents the indicator function. The consistency over a set of prompts may be aggregated into a global consistency score, $\gC$, by averaging all the conditioning-wise DSG scores, $\gC^{p}$.

\noindent\textbf{Conditional diversity, $\gD_C$.} 
We measure per-prompt conditional diversity as follows:
\begin{equation}
    \gD_C^{p} = \frac{1}{N^2 - N}\sum_{j\neq i}{\gS(\enc(\tY_{j}), \enc(\tY_{i}))},
    \label{eq:cond_div}
\end{equation}
where $\gS$ is a similarity or distance function, and $\enc$ is an image feature extractor. In our analysis, we use cosine similarity and the DreamSim~\citep{fu2023dreamsim} feature extractor. DreamSim leverages an ensemble of modern vision encoders, including DINO~\citep{caron2021emerging} and two independently trained CLIP encoders, and is reported to correlate well with human perception. The conditional diversity over a set of prompts may be aggregated into a global score, $\gD_C$, by averaging all the conditioning-wise scores, $\gD_C^{p}$.\looseness-1

\noindent\textbf{Conditional realism, $\gR_C$.} 
We measure per-prompt conditional realism as follows:
\begin{equation}
    \gR^{p}_C = \frac{1}{N}\sum_{j=1}^{N}{\max_{i} (\gS(\enc(\tX_i), \enc(\tY_{j}))}), \quad i \in \{1, \dots, N^\prime\},
    \label{eq:cond_real}
\end{equation}
where $\tX \in \mathbbm{R}^{N^\prime\times H\times W\times 3}$ represents a tensor of $N^\prime$ real images. Note that both $\tX$ and $\tY$ represent generations and real images of the same prompt $\rvp$. Similarly to conditional diversity, we implement $\gS$ as  cosine similarity and use DreamSim as feature extractor. The conditional realism over a set of prompts may be aggregated into a global score, $\gR_C$, by averaging all the conditioning-wise scores $\gR^{p}_C$.\looseness-1

\noindent\textbf{Marginal diversity, $\gD_M$.} Commonly used metrics of marginal diversity, such as \emph{recall}~\citep{sajjadi2018assessing,kynkaanniemi2019improved} or \emph{coverage}~\citep{naeem2020reliable}, compare real and generated image distributions by leveraging a reference dataset of real images to ground the notion of diversity. Marginal diversity may also be measured with metrics which do not rely on a reference dataset, such as the Vendi Score~\citep{friedman2023vendi}. In our analysis, we use recall~\citep{sajjadi2018assessing,kynkaanniemi2019improved} to compute marginal diversity given its ubiquitous use in the literature. Recall measures marginal diversity as the probability that a random real image falls within the support of the generated image distribution.\looseness-1

\noindent\textbf{Marginal realism, $\gR_M$.} The most commonly used metric to estimate image realism is the Fréchet Inception Distance (FID)~\citep{heusel2017gans}. FID relies on a pre-trained image encoder (usually, the Inception-v3 model trained on ImageNet-1k~\citep{szegedy2015going}) that embeds both generated and real images from a reference dataset. The metric estimates the distance between distributions of features of real images and features of generated images, relying on a Gaussian distribution assumption. The FID summarizes image quality and diversity into a single scalar. In our analysis, to disentangle both axes of evaluation, we use precision~\citep{kynkaanniemi2019improved, naeem2020reliable} as marginal realism metric. Precision measures marginal realism as the probability that a random generated image falls within the support of the real image distribution.\looseness-1

\subsection{\Cdq knobs}
\label{sec:knobs}

\noindent\textbf{Guidance scale.}
To control the strength of the conditioning, a guidance scale (\kscale)  hyper-parameter can be used to bias the  sampling of diffusion models like DDPM~\citep{ho2020denoising}, see \eg, classifier~\citep{dhariwal2021diffusion} or classifier-free guidance (CFG) ~\citep{ho2021classifierfree}. 
More precisely, rewriting \cref{eq:cond_gen} for diffusion models trained with CFG, we obtain:
\begin{equation}
    \tY = \lambda \gen(\tZ, \rvp) + (1-\lambda)\gen(\tZ,\emptyset),
\end{equation}
where $\lambda$ is the guidance scale, $\emptyset$ is an empty conditioning prompt, and the first and second terms indicate conditional and unconditional samplings, respectively. Importantly, $\lambda$ can be arbitrarily increased in order to steer the model to generate samples more aligned with the conditioning $\rvp$.

\noindent\textbf{Post-hoc filtering.} 
To improve the generated images, 
\eg in terms of realism or consistency, or to avoid certain undesirable  generations, a set of  images generated for the same prompt may be filtered to retain the  top-$m$ images based on a predefined criterion, which can be either based on human preferences or automatic metrics. Considering the latter case, a common choice of metric is the CLIPScore, resulting in:
\begin{equation}
    \tY = \topk\biggr(m, \ \gS(\rvp, \enc(\tY_{j}))\biggr),
\end{equation}
where decreasing $m$ ensures higher consistency.\looseness-1

\noindent\textbf{Retrieval-augmented generation.}
Generation can be conditioned on additional information, \eg via nearest-neighbor search in a database given a query image. 
\begin{equation}
    \tY = \gen(\tZ, \oplus_{\rvp_j \in \sK \cup \{\rvp\}}),%
    \label{eq:cond_gen2}
\end{equation}
where $\oplus$ denotes the aggregation operator and $\sK$ is the set of nearest neighbors of $\rvp$.
Existing retrieval-augmented \igenmodels adopt different aggregation operators. For instance, RDM~\citep{blattmann2022semi}, KNN-Diffusion~\citep{sheynin2023knndiffusion}, and Re-Imagen~\citep{chen2022re}, concatenate the retrieved vectors, and use cross-attention to condition the generative process. Autoregressive models like RA-CM3~\citep{yasunaga2023Retrieval} and CM3Leon~\citep{yu2023scaling}, concatenate the retrieved vectors to the input before performing self-attention. Regardless of the type of aggregation, changing the value of $k$ in retrieval-augmented models can affect the conditional diversity and consistency of the generations.

\noindent\textbf{Compression rate.}%
Neural image compression models are generative autoencoder-like models that learn to compress images into low-dimensional representations before reconstructing them.  The compression rate, usually expressed in bits-per-pixel (bpp), determines the ability to faithfully reconstruct the original image. %
Some neural image compression models, such as PerCo~\citep{careil2024towards}, treat compression as a conditional generative modeling problem, allowing to sample approximate reconstructions given the compressed image code.
In such cases, we could expect that by reducing the bitrate, the model might trade consistency/realism for conditional diversity  as the compressed image code will carry less information about the original image. 

\subsection{Pareto fronts}

We perform an optimization over state-of-the-art models and their knobs with the goal of capturing the \cd, \qd, and \cq Pareto fronts that are currently reachable, and building understanding on the \cdq multi-objective. More precisely, we quantify consistency, diversity and realism for each model-knob-value pair with the metrics presented in Section~\ref{sec:metrics}. We then leverage all the resulting measurements to obtain the Pareto fronts that capture optimal the \cdq values achieved by current state-of-the-art \cigenmodels. For visualization ease, we transform the multi-objective into three bi-objectives: \cd, \qd and \cq.\looseness-1

\section{Experiments}
\label{sec:experiments}

In this section, we study T2I and I-T2I models and depict the achievable \cdq Pareto fronts by altering the models and their associated knobs. We start by detailing the experimental setups and follow with a detailed discussion of results, covering T2I models in Section~\ref{sec:T2Iparetos} and I-T2I models in Section~\ref{sec:I-T2Iparetos}. We then highlight the utility of our approach in a geodiversity analysis in Section~\ref{sec:GeoDEParetos}. Finally, we study the impact of using different knobs to control these tradeoffs in Section~\ref{sec:knobParetos}.\looseness-1

\begin{table}
\centering
\caption{Knobs for text-to-image (T2I) and text\&image-to-image (I-T2I) models used in our study. For \rdm, 1.3M corresponds to the training dataset, while 20M to the retrieval database.} %
\label{tab:models}
{
\small
\begin{tabular}{lcccccc}
\toprule
\multirow{2}{*}{\textbf{Model}} & \multirow{2}{*}{\textbf{Dataset size}} & \multicolumn{4}{c}{\textbf{Knobs}} \\ \cmidrule(lr){3-6} 
 &  & \kscale & \ktopk & \kneighbor & \kcomp \\
 \midrule
\multicolumn{6}{l}{\textit{T2I}} \\
\sdone~\citep{rombach2022high} & $\sim$2B & \yes & \yes & \no & \no \\
\sdtwo~\citep{rombach2022high} & $\sim$2B & \yes & \yes & \no & \no \\
\sdxl~\citep{podell2023sdxl} & $\sim$2B & \yes & \yes & \no & \no \\
\sdxlt~\citep{sauer2023adversarial} & $\sim$2B & \no & \yes & \no & \no \\
\midrule
\multicolumn{6}{l}{\textit{I-T2I}} \\
\perco~\citep{careil2024towards} & $\sim$300M & \yes & \yes & \no & \yes \\
\rdm~\citep{blattmann2022semi} & 1.3M + 20M & \yes & \yes & \yes & \no \\
\unclip~\citep{ramesh2022hierarchical} & $\sim$2B & \yes & \yes & \no &  \no \\
\bottomrule
\end{tabular}%
}
\end{table}

\noindent\textbf{Models.}
We consider different state-of-the-art \cigenmodels and group them by their conditioning modalities. %
For T2I models, we consider several versions of LDM: \sdone, \sdtwo~\citep{rombach2022high}, \sdxl~\citep{podell2023sdxl}\footnote{For \sdxl we use the base model v1.0 without the refiner}, and \sdxlt~\citep{sauer2023adversarial}. 
For I-T2I models, we pick \unclip~\citep{ramesh2022hierarchical}, \rdm~\citep{blattmann2022semi}, and the neural image compression model  \perco~\citep{careil2024towards}, which conditions an LDM with a quantized image representation together with its caption\footnote{We note that \perco usually caption the input image with a captioner, while in our case we get the caption from the dataset.\looseness-1}. We summarize the models considered in our analysis and their knobs in \cref{tab:models}.\looseness-1

\noindent\textbf{Datasets.} 
We benchmark the models on a popular computer vision dataset, MSCOCO~\citep{lin2014microsoft, caesar2018coco}. %
In particular, we use the validation set from the 2014 split~\citep{lin2014microsoft}, which contains 41K images, to compute the marginal metrics, and the 2017 split~\citep{caesar2018coco}, which contains 5K images, to compute the conditional metrics. This choice is mostly to limit computational costs, as conditional metrics require multiple samples for each conditioning. In addition, we benchmark geographic representation with GeoDE \citep{ramaswamy2024geode}, which contains images from everyday objects in countries across six geographic regions. Following \citep{hall2024dig}, we balance the dataset across 27 objects, yielding 29K images and 162 unique \texttt{\{object\} in \{region\}} prompts.

\noindent\textbf{Implementation details.}
We adopt the \texttt{Diffusers} library for the LDM models~\citep{vonplaten2022diffusers} and the official models' repos for \rdm and \perco. We set the number of inference steps to 50 (20 for \perco as suggested in their paper) using deterministic sampling strategies, DPM++~\citep{lu2022dpm} for \texttt{Diffusers} models and DDIM~\citep{song2020denoising} for others. For the conditional metrics on MSCOCO, we sample $10$ images per prompt, using the 5,000 image-caption pairs of the 2017 validation split,  while for the marginal metrics we sample $1$ image per conditioning, using 30,000 randomly selected data points from the validation set of 2014. Note that, as MSCOCO contains multiples captions for each image, we fix the first caption as prompt for generations.
For GeoDE, we sample $180$ images for each of the \texttt{\{object\} in \{region\}} prompts for both conditional and marginal metrics. We disaggregate metrics by groups, per \citet{hall2024dig}, to measure disparities between geographic regions. 
For metrics based on \dreamsim we use the ensemble backbone as recommended from the official repository%
. For marginal metrics we use the implementation of \texttt{prdc}%
.
For DSG%
, we leverage \texttt{GPT-3.5-turbo} to generate questions from the prompts, and \texttt{InstructBLIP}~\citep{dai2024instructblip} to make the predictions. When performing \ktopk based on \clipscore, we use \texttt{CLIP-ViT-H-14-s32B-b79K} from Hugging Face. %
Finally, we ablate different values for each knob as reported in \cref{suppl:details}.\looseness-1

\begin{figure*}[ht]
    \centering
    \includegraphics[height=.55cm]{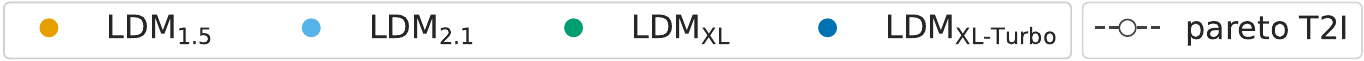}
    \includegraphics[width=\textwidth]{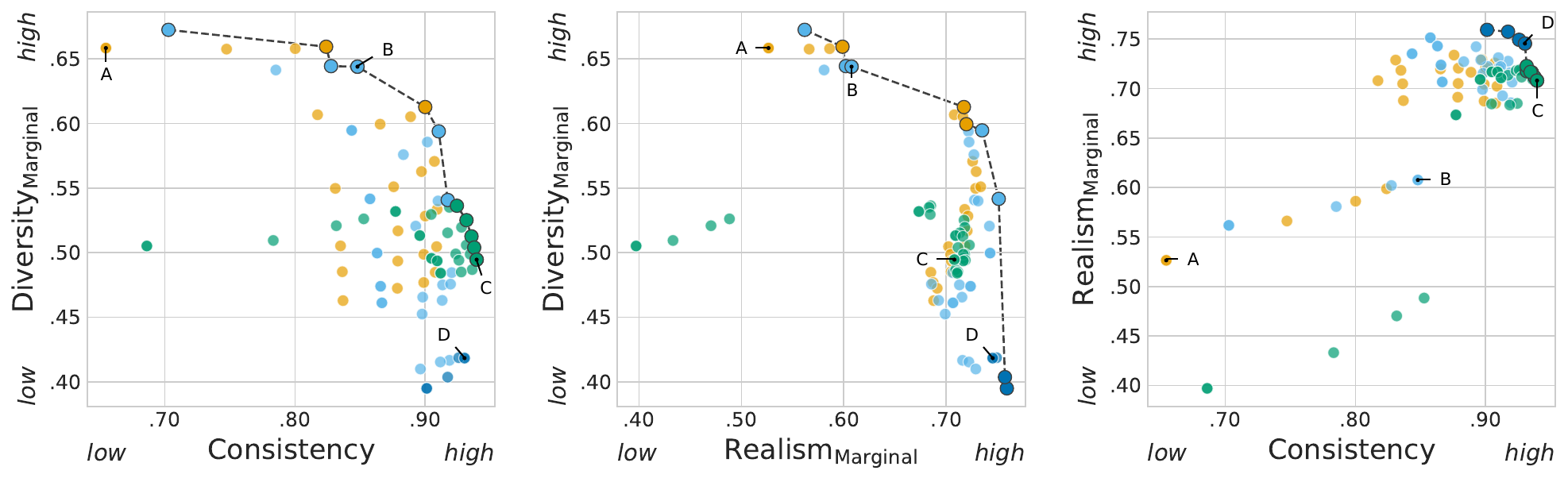}
    \includegraphics[width=\textwidth]{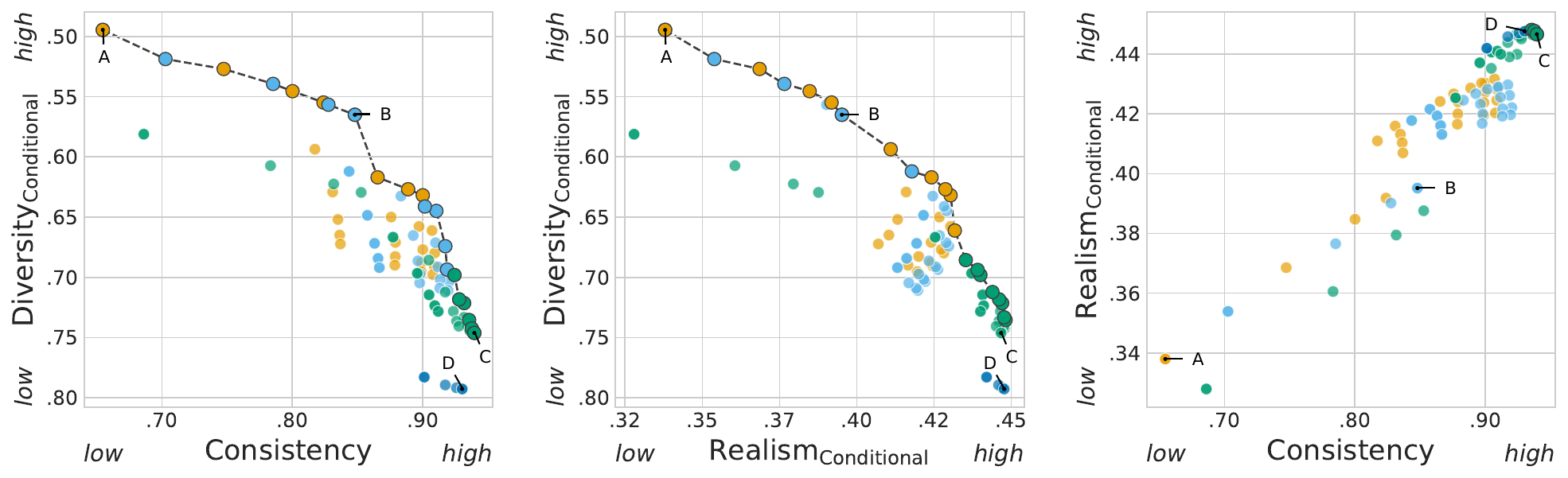}
    \caption{\Cd, \qd and \cq Pareto fronts for T2I generative models.  (top) marginal, (bottom) conditional metrics. Each dot is a configuration of model's knobs. Labeled dots (A-D) are visualized in \cref{fig:quali_t2i_pareto}. \looseness-1}
    \label{fig:paretoT2I}
\end{figure*}

\begin{figure*}[ht]
    \centering
    \begin{subfigure}[b]{.49\textwidth}
    \includegraphics[height=6.8cm]{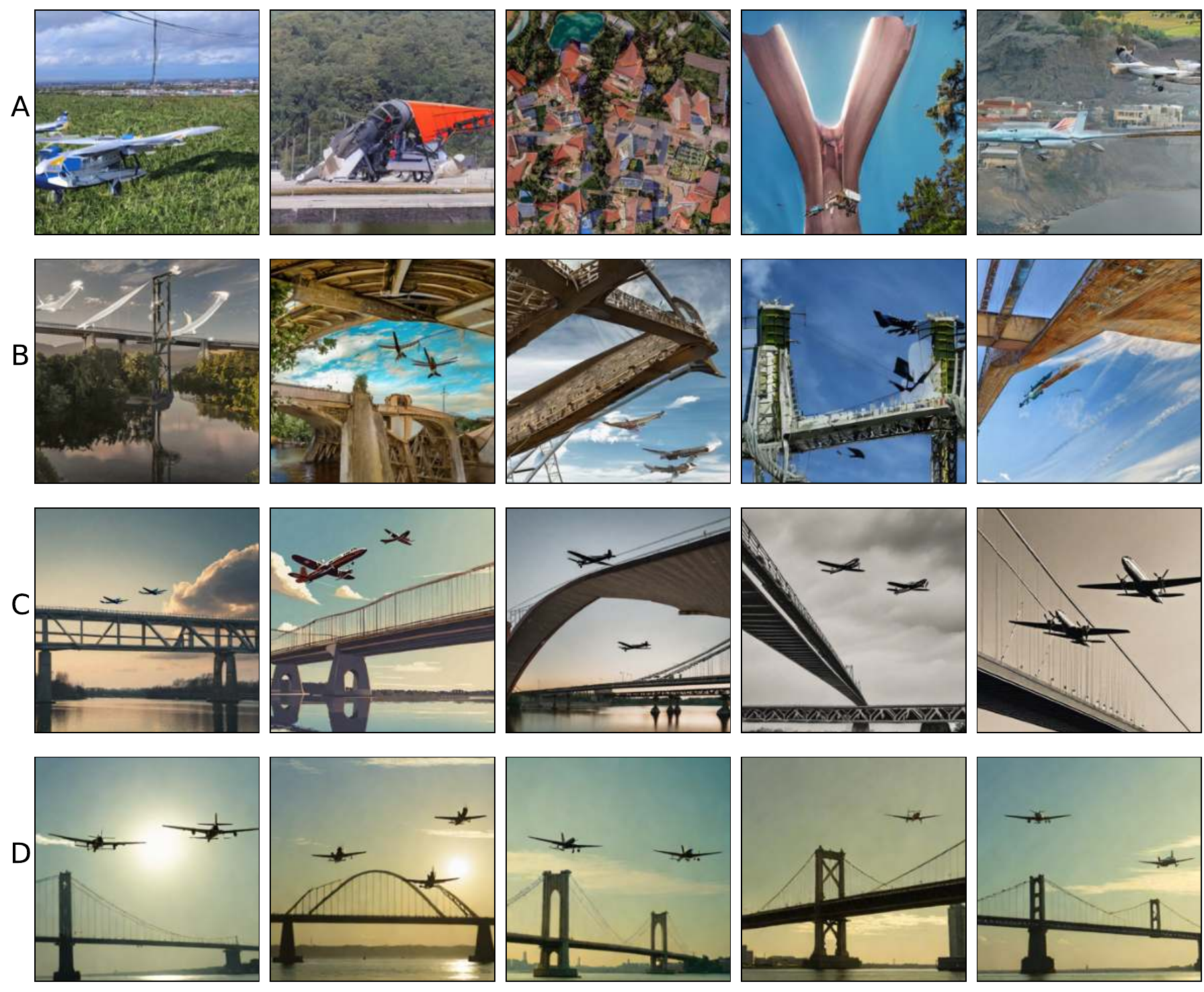}
    \end{subfigure}
    \hfill
    \begin{subfigure}[b]{.49\textwidth}
    \includegraphics[height=6.8cm]{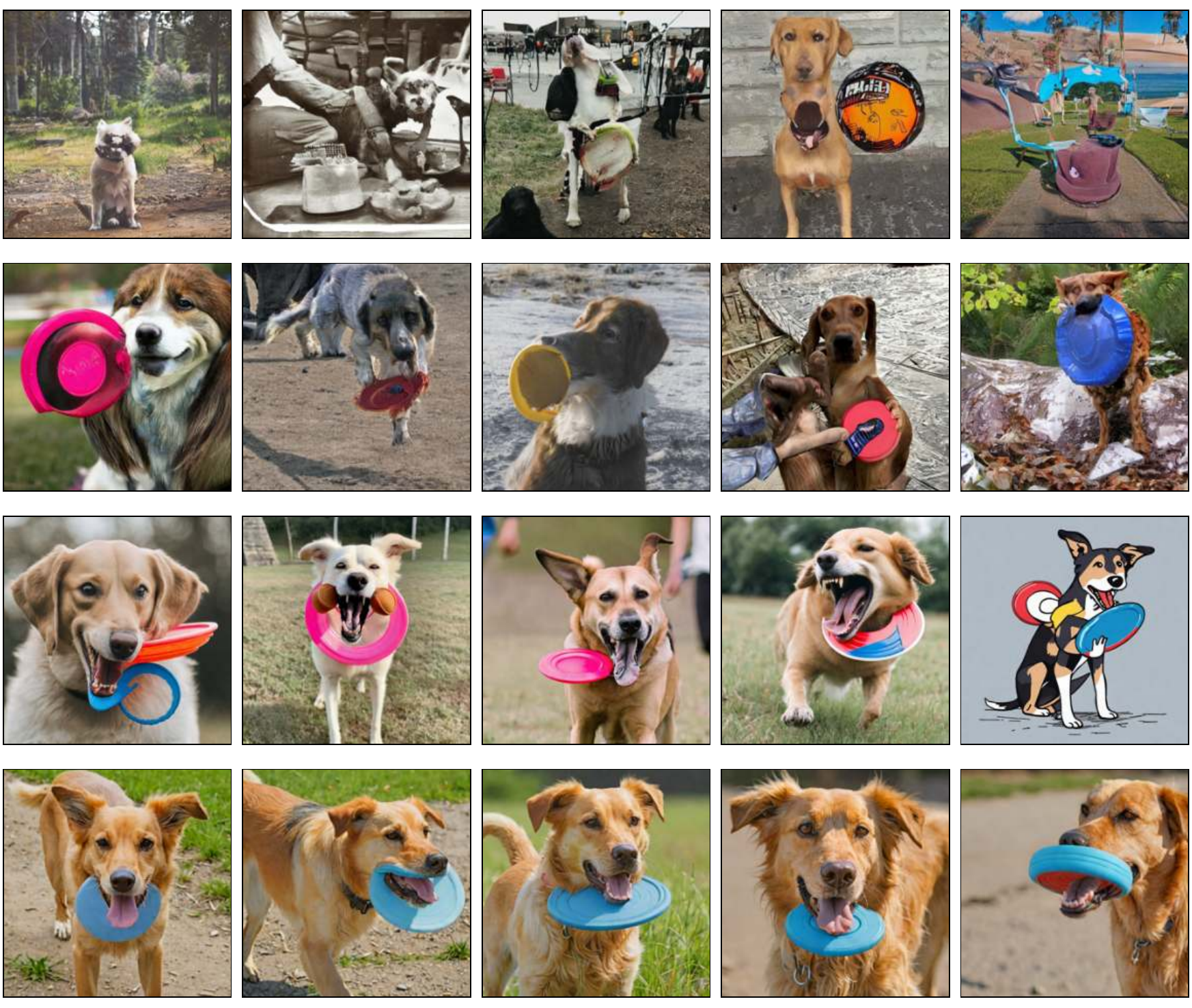}
    \end{subfigure}
    \caption{T2I qualitative results on \coco. A-D refer to the models marked in \cref{fig:paretoT2I}. (left) \texttt{Two planes flying in the sky over a bridge.} (right) \texttt{There is a dog holding a Frisbee in its mouth.}}
    \label{fig:quali_t2i_pareto}
\end{figure*}

\subsection{\Cdq multi-objective for text-to-image models}
\label{sec:T2Iparetos}

In ~\cref{fig:paretoT2I}, we depict \cd, \qd and \cq Pareto fronts for open source T2I generative models. In particular, \cref{fig:paretoT2I}~(top) depicts marginal realism and diversity metrics while \cref{fig:paretoT2I}~(bottom) shows their conditional counterparts. Note that consistency is computed in the same way (\dsg) in both figures. We now discuss each of the pair-wise metrics Pareto fronts.\looseness-1

\noindent\textbf{\Cd.} The Pareto fronts 
in \cref{fig:paretoT2I} (left, top and bottom),
are composed of three models: \sdone, \sdtwo and \sdxl. We observe that improvement in diversity, both marginal (Recall) and conditional  (\dreamsim score), comes at the expense of consistency (\dsg). On the one hand, \sdtwo and \sdone achieve the best marginal and conditional diversities, respectively. On the other hand, and perhaps unsurprisingly, \sdxl reaches the best consistency ( $\geq$~$95$\% of DSG accuracy), while \sdone and \sdtwo dominate the middle region of the frontier. Moreover, by comparing these two models, we notice that Pareto optimal hyperparameter configurations of \sdtwo obtain slightly higher consistency scores. In \cref{fig:quali_t2i_pareto}, we validate these observations showcasing samples from \sdone (A) at high-diversity/low-consistency, \sdtwo (B) from the middle of the frontier, and \sdxl (C) at high-consistency/low-diversity. Both in the case of the ``two planes'' and of the ``dog'', the variance of colors and backgrounds are reduced when visual quality is increased. Other samples are in \cref{suppl:results}.~\looseness-1

\noindent\textbf{\Qd.}
The marginal \qd (Precision-Recall) Pareto front in \cref{fig:paretoT2I}~(middle, top),
is composed of three models: \sdone, \sdtwo and \sdxlt. In this case, we also observe a tradeoff:  higher marginal diversity coincides with lower realism for \sdone and \sdtwo. \sdxlt obtains the samples of highest realism. However, we observe that the realism gain compared to \sdtwo is rather small and leads to a steep decrease in sample diversity. We attribute this drop to the  adversarial objective used to distill \sdxlt from \sdxl, as also noted by \citet{sauer2023adversarial}.
Interestingly, %
\sdxl does not appear on the Pareto front, and it is even quite far away from it. This is probably due to \sdxl (without refiner) generating smooth images lacking of high frequency details (\eg, see the dog in \cref{fig:quali_t2i_pareto} and \citep{podell2023sdxl}), and the marginal metrics, which are computed with InceptionV3 features, are sensitive to those frequencies~\citep{geirhos2018imagenet}. 
Instead, by looking at the conditional metrics in \cref{fig:paretoT2I} (middle), which are based on \dreamsim that extract more sematical features~\citep{fu2023dreamsim}, we observe that \sdxl belongs to the Pareto front together with \sdone, \sdtwo. In particular, \sdxl achieves the best conditional realism,  obtained at the expense of conditional diversity. Here, we remark that \sdxlt only gets comparable (slightly lower) realism but considerably lower diversity. This difference is evident by looking at C (\sdxl) vs. D (\sdxlt) in \cref{fig:quali_t2i_pareto}. When comparing Pareto optimal points of \sdone and \sdtwo, we note that \sdone reaches slightly better conditional realism than \sdtwo.

\noindent\textbf{\Cq.}
In  \cref{fig:paretoT2I} (right, top and bottom) 
we observe that realism and consistency show relatively strong positive correlation as improvement in one metric oftentimes leads to an improvement in the other metric, with the correlation being stronger for the conditional metrics than for the marginal ones. We observe that the Pareto front is dominated by \sdxl and \sdxlt model, highlighting how the advancement of T2I generative models have favored \cq over the diversity objective. Indeed, we can also notice that in the distribution of non-Pareto-optimal points, \sdtwo seems better than \sdone, matching the historical development of these models. 

\begin{tcolorbox}[my_style]
\subsubsection*{Key insights}
\begin{itemize}[leftmargin=*]
    \item Progress in T2I models has been driven by improvements in realism and/or consistency. State-of-the art T2I models improve consistency and/or realism by sacrificing representation diversity. Yet, improvements in realism are correlated with improvements in consistency.%
    \item More recent models should be used when the downstream task requires samples with high realism  -- \sdxlt -- and consistency -- \sdxl --. However, older models -- \sdone and \sdtwo -- are preferable for tasks that require good representation diversity. 
    \item Both marginal and conditional metrics display correlated Pareto fronts. 
\end{itemize}
\end{tcolorbox}

\begin{figure*}[ht]
    \centering
    \includegraphics[height=.55cm]{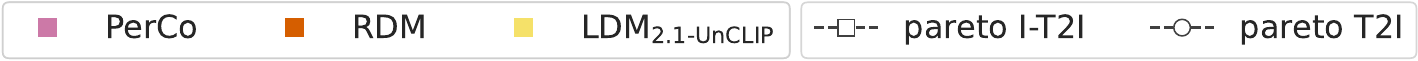}
    \includegraphics[width=\textwidth]{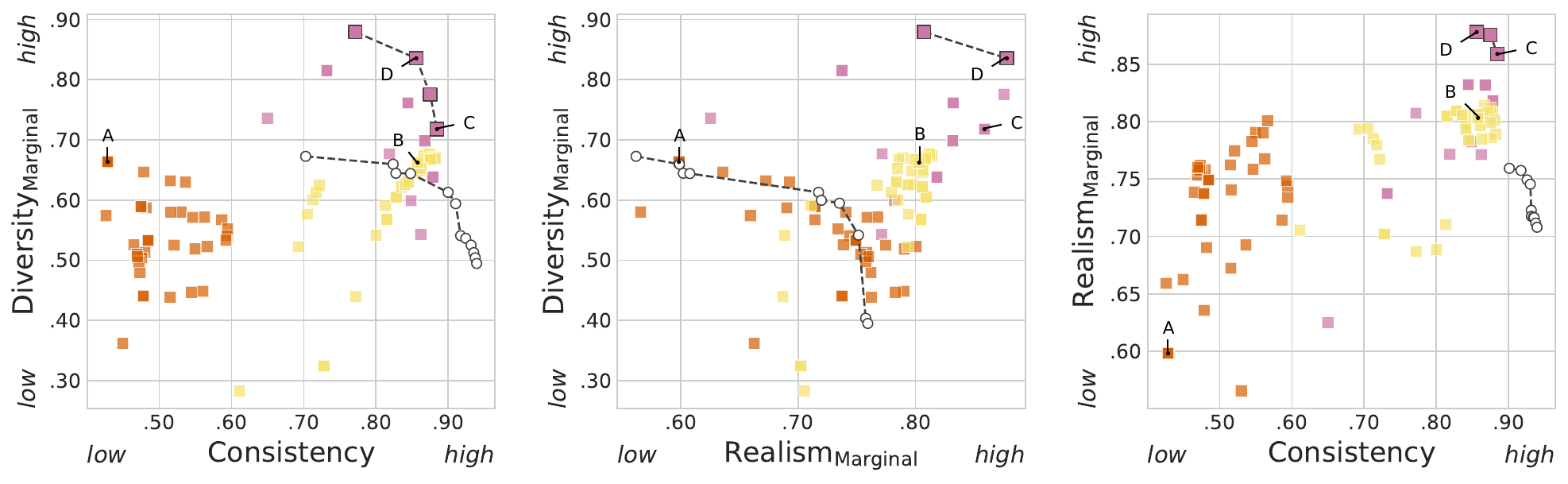}
    \includegraphics[width=\textwidth]{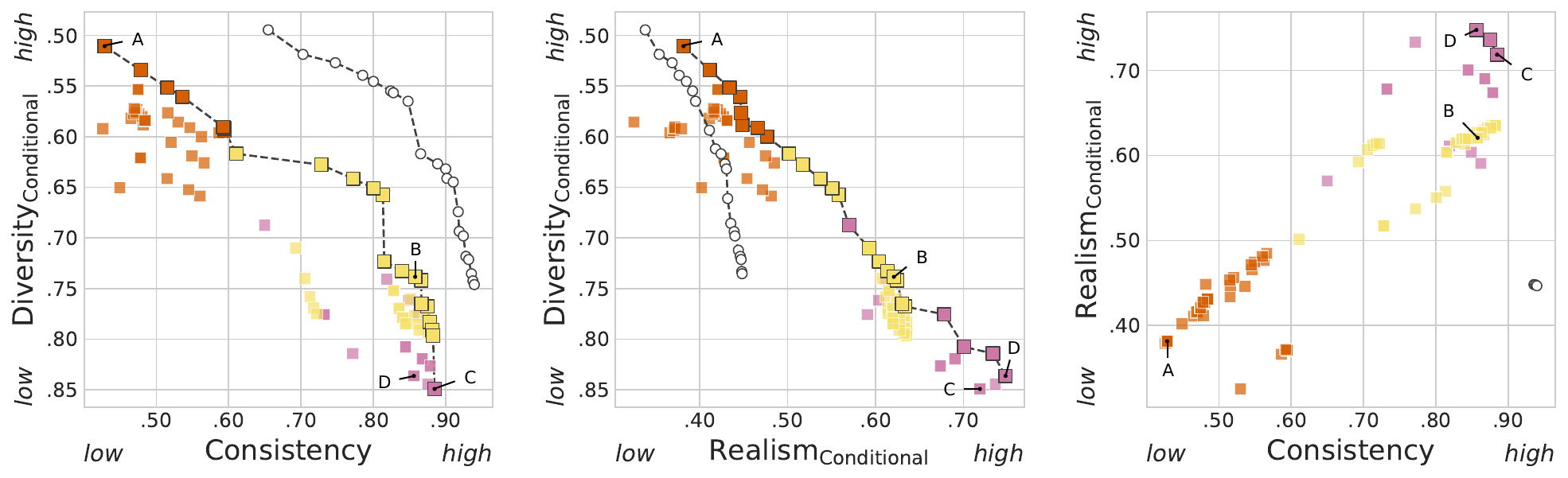}
    \caption{ \Cd, \qd and \cq Pareto fronts for I2I and I-T2I generative models. (top) marginal, (bottom) conditional metrics. Each dot is a configuration of model's knobs. Labeled dots are visualized in \cref{fig:quali_it2i_pareto}\looseness-1}
    \label{fig:paretoI-T2I}
\end{figure*}

\begin{figure*}[ht]
    \centering
    \begin{subfigure}[b]{.49\textwidth}
    \includegraphics[height=6.8cm]{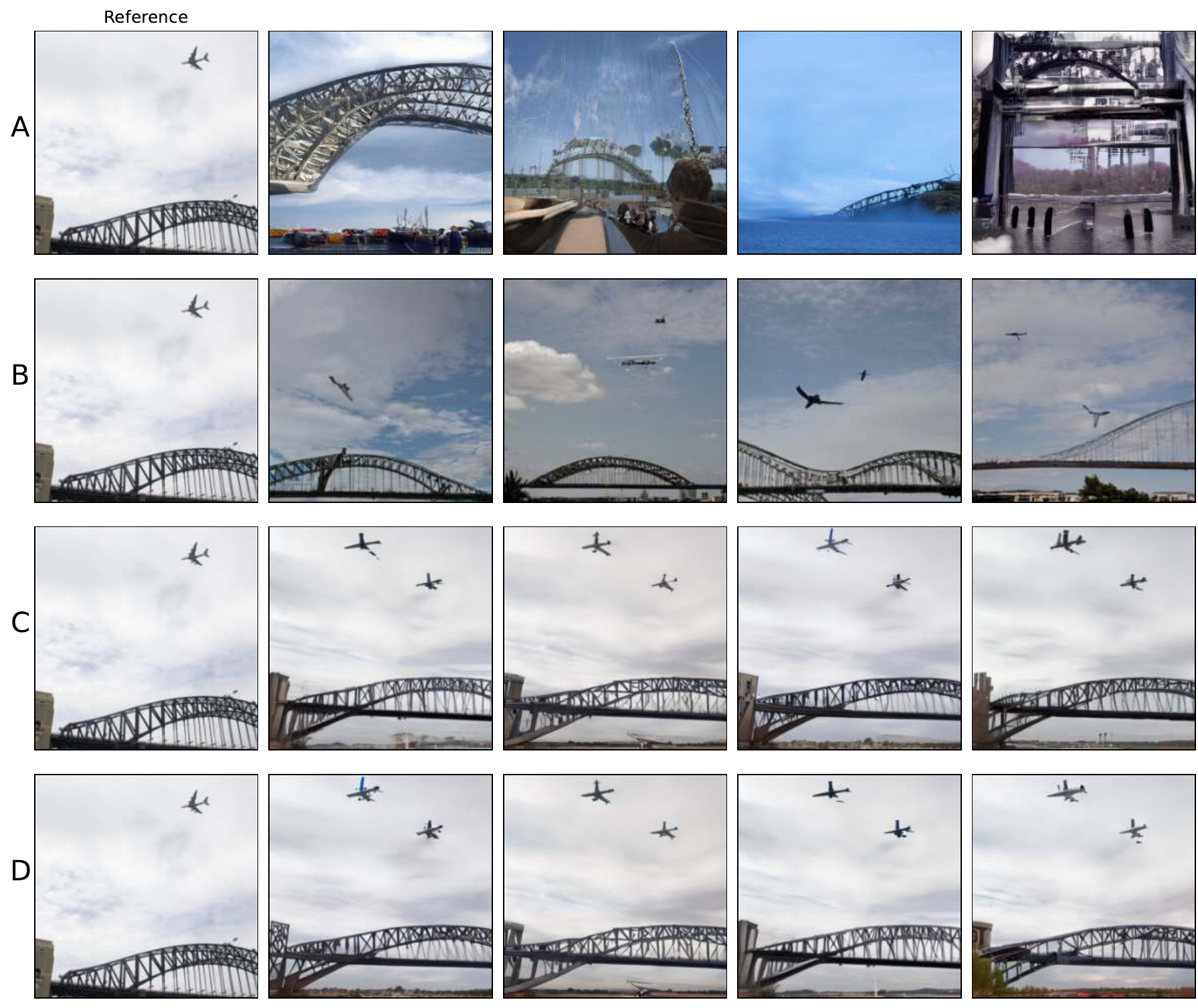}
    \end{subfigure}
    \begin{subfigure}[b]{.49\textwidth}
    \includegraphics[height=6.8cm]{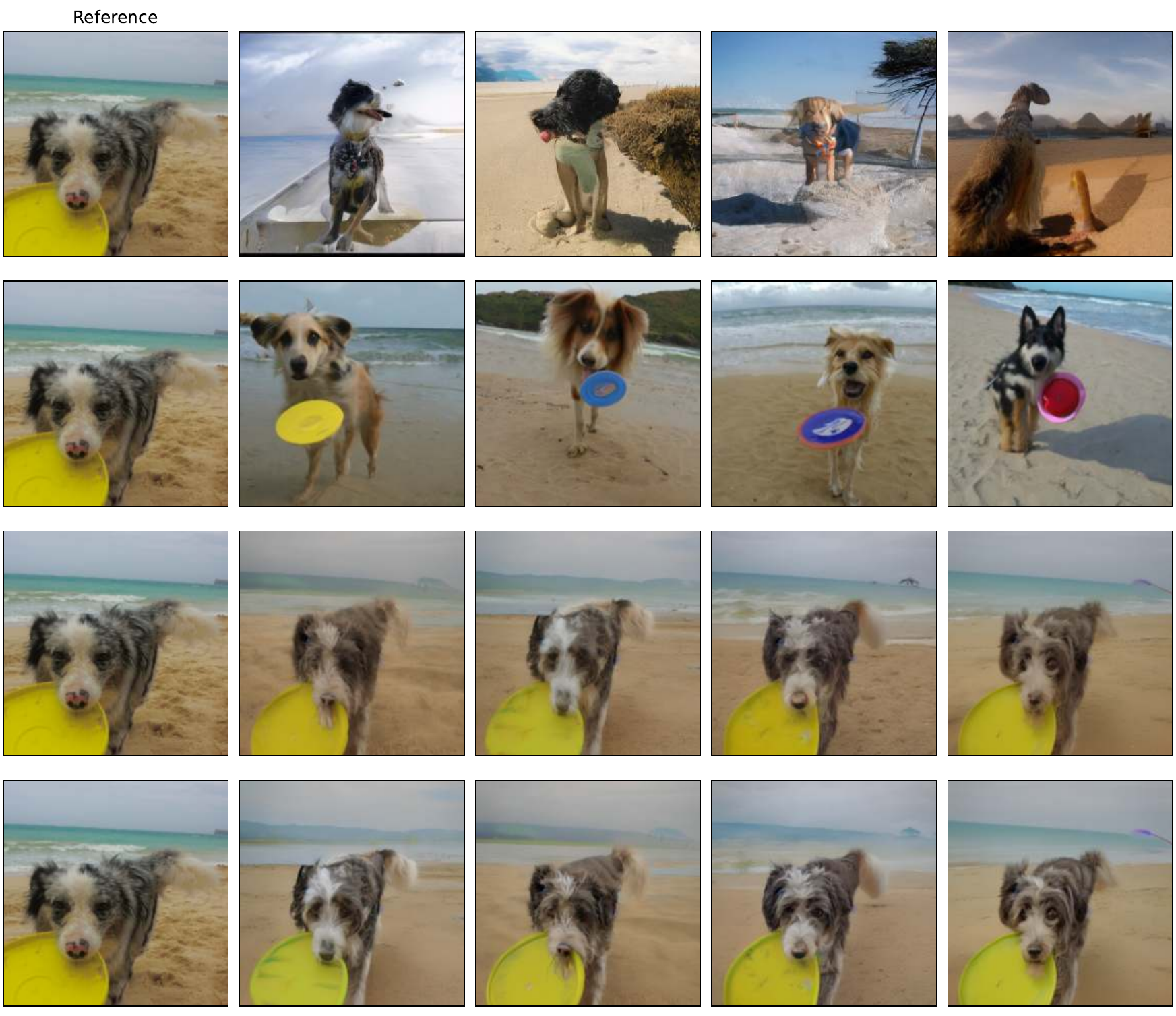}
    \end{subfigure}
    \caption{I-T2I qualitative results on \coco. A-D refer to the models marked in \cref{fig:paretoI-T2I}. ``Reference'' column shows the conditioning image. \todo{fix cropping in reference, consider dropping last row}}
    \label{fig:quali_it2i_pareto}
\end{figure*}

\subsection{Pareto fronts of image\&text-to-image models}
\label{sec:I-T2Iparetos}

\noindent\textbf{\Cd.}
The marginal \cd Pareto front in \cref{fig:paretoI-T2I} (left, top) does not show a clear tradeoff, as it is only composed by \perco neural compression models achieving both high consistency and  diversity. 
On the contrary, for the conditional metrics (left, bottom), the tradeoff is clearly noticeable. The Pareto front is composed of three models: \rdm, \unclip, and \perco, \rdm reaching the best conditional diversity, \unclip populating a large portion -- from mid to high consistency -- of the tradeoff, and \perco achieving the highest consistency, but only for a small margin. We visualize samples from these models in \cref{fig:quali_it2i_pareto} (A,B,C, respectively), confirming the findings
exposed by the metrics. It is important to note that \perco achieves the highest marginal diversity and the lowest conditional diversity; this is expected given the goal of a compression model to yield good reconstructions of the data. Obtaining high realism reconstructions allows for a good reconstruction of the real data manifold, which in turn results in high recall. However, in this case, multiple reconstructions of the same image will all look very similar, hence producing low conditional diversity.\todo{comment on the two paretos}\looseness-1

\noindent\textbf{\Qd.}
 Considering marginal metrics in \cref{fig:paretoI-T2I} (middle, top), \perco is again the only model producing Pareto optimal points, with even higher realism (precision) and diversity (recall) scores. Also, the non-Pareto optimal points are mostly disposed along the main diagonal of the plot, suggesting rather small \qd compromises. Instead, and once again similarly to the \cd case, the Pareto fronts obtained from conditional diversity and realism (middle, bottom) contain all the three models considered, with \rdm model producing the most conditionally diverse samples and \perco producing the samples with the highest conditional realism. Thus, by optimizing the model towards conditional realism, the conditional diversity is being sacrificed.  

\noindent\textbf{\Cq.} Similarly to T2I models, in \cref{fig:paretoI-T2I} (right, top and bottom) we observe a correlation between realism (marginal or conditional) and consistency. Perhaps unsurprisingly, \perco achieves the best results in terms of both realism and consistency, and is the only model producing Pareto optimal points. Despite not making it to the Pareto, we can still compare \unclip and \rdm as their hyperparameter configurations constitute two easily separable clusters, with \rdm achieving much lower consistency and realism than the worst hyperparameter configuration of \unclip. We might attribute this difference to the different dataset scale (millions vs billions) and model capacities (400M vs. 840M) of the \rdm and \unclip, respectively.

\begin{tcolorbox}[my_style]
\subsubsection*{Key insights}
\begin{itemize}[leftmargin=.3em]
    \item Progress in I-T2I models has been driven by improvements in realism and/or consistency. State-of-the-art models often trade realism for conditional diversity, this tradeoff is not visible when considering marginal diversity.
    \item Marginal diversity is dominated by the models that reconstruct more faithfully the conditioning image. This is not the case for the conditional metric that is sensitive to conditional diversity that oftentimes is important in downstream applications.
    \item Compression models, \perco, should be prioritized when working on downstream applications that require high realism and consistency. However, when the downstream application requires high representation diversity, \rdm and \unclip are preferable.
\end{itemize}
\end{tcolorbox}

\subsection{Pareto fronts for geographic disparities in T2I models}
\label{sec:GeoDEParetos}
\begin{figure}[ht]
    \centering
    \includegraphics[height=.9cm]{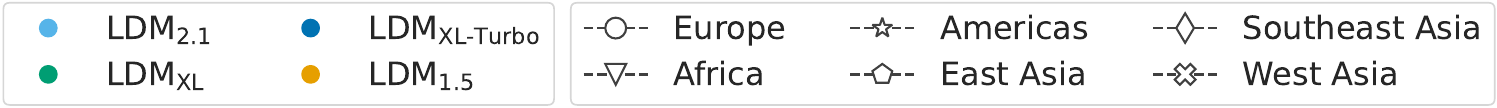}
    \includegraphics[width=\textwidth]{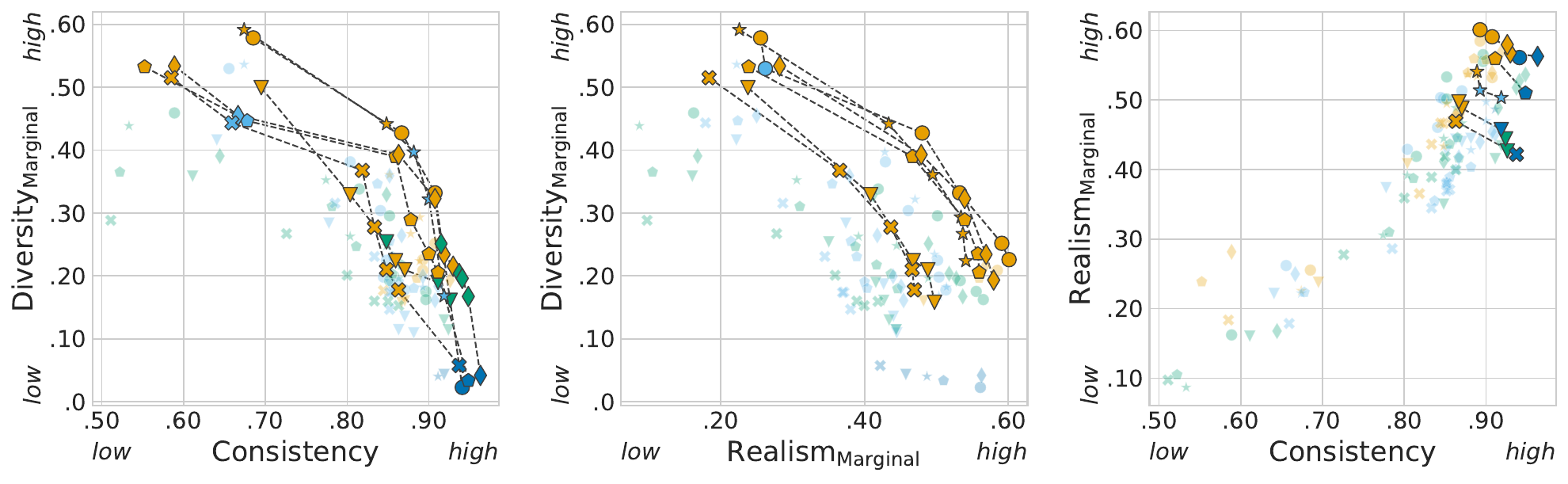}
    \includegraphics[width=\textwidth]{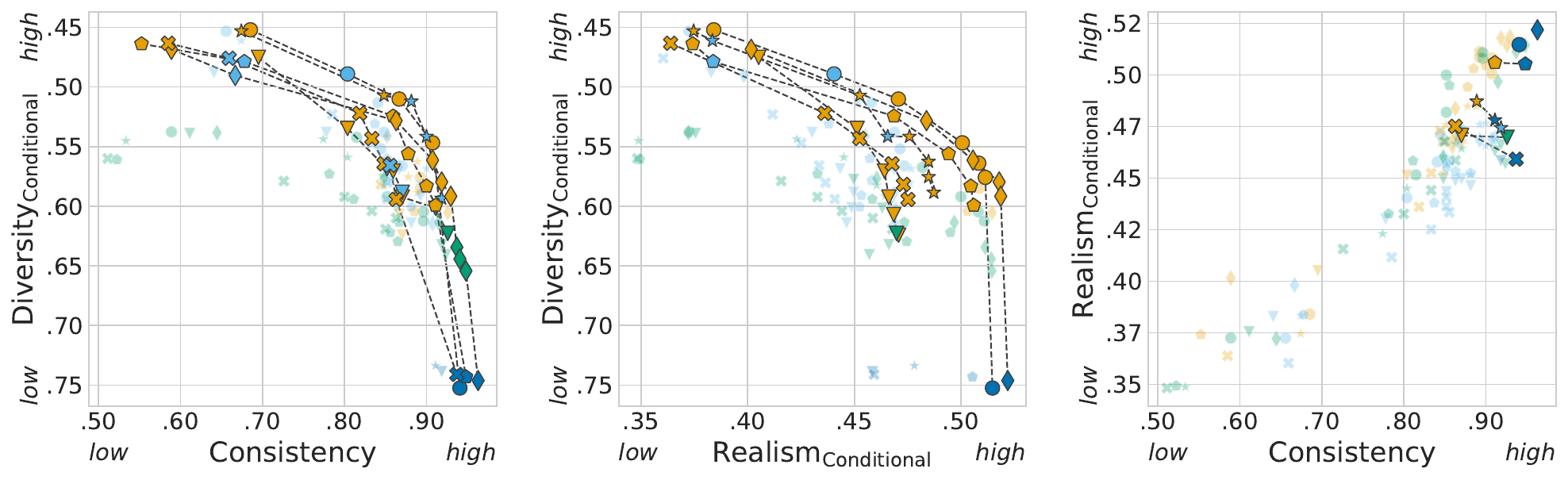}
    \caption{\Cd, \qd, and \cq Pareto fronts for T2I models on the GeoDE dataset. Consistency measures only the presence of the object in the image. Each models' configuration differ solely for guidance scale value.
    }
    \label{fig:pareto_geode}
\end{figure}

\begin{figure*}
    \includegraphics[width=.49\textwidth]{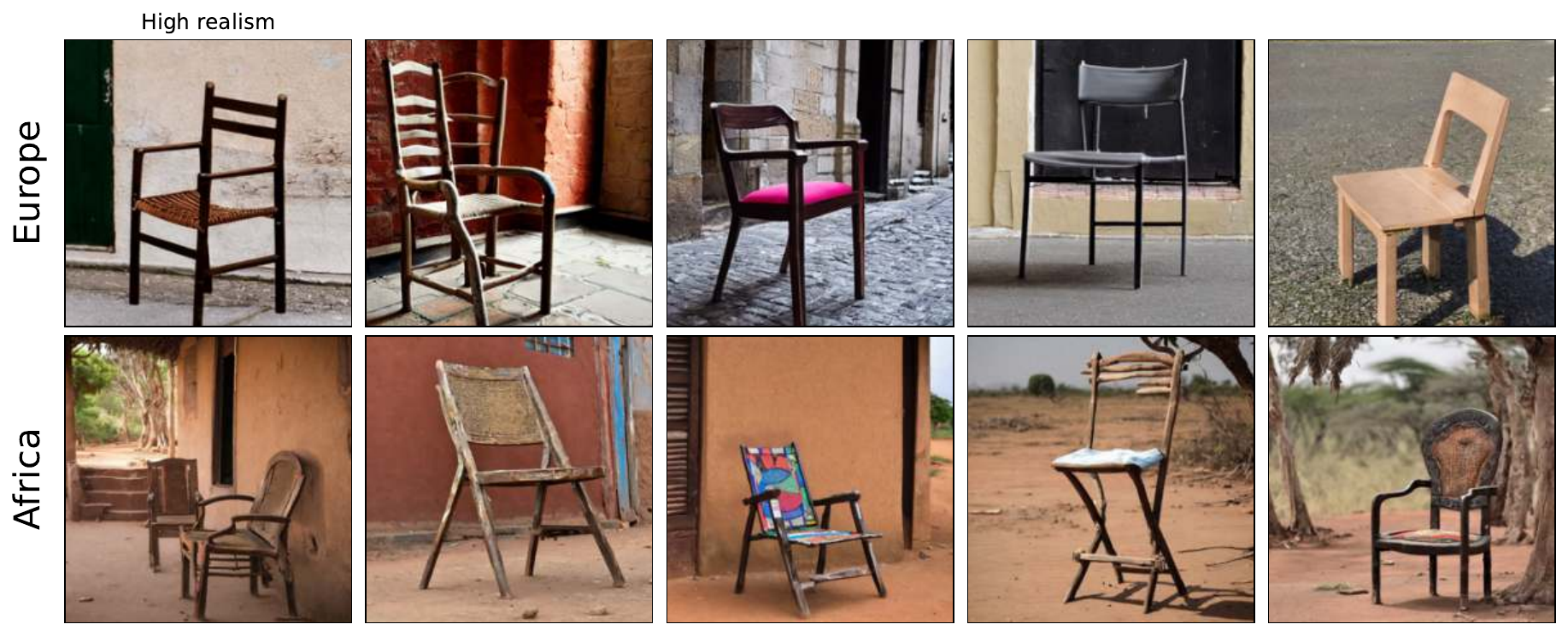}
    \includegraphics[width=.49\textwidth]{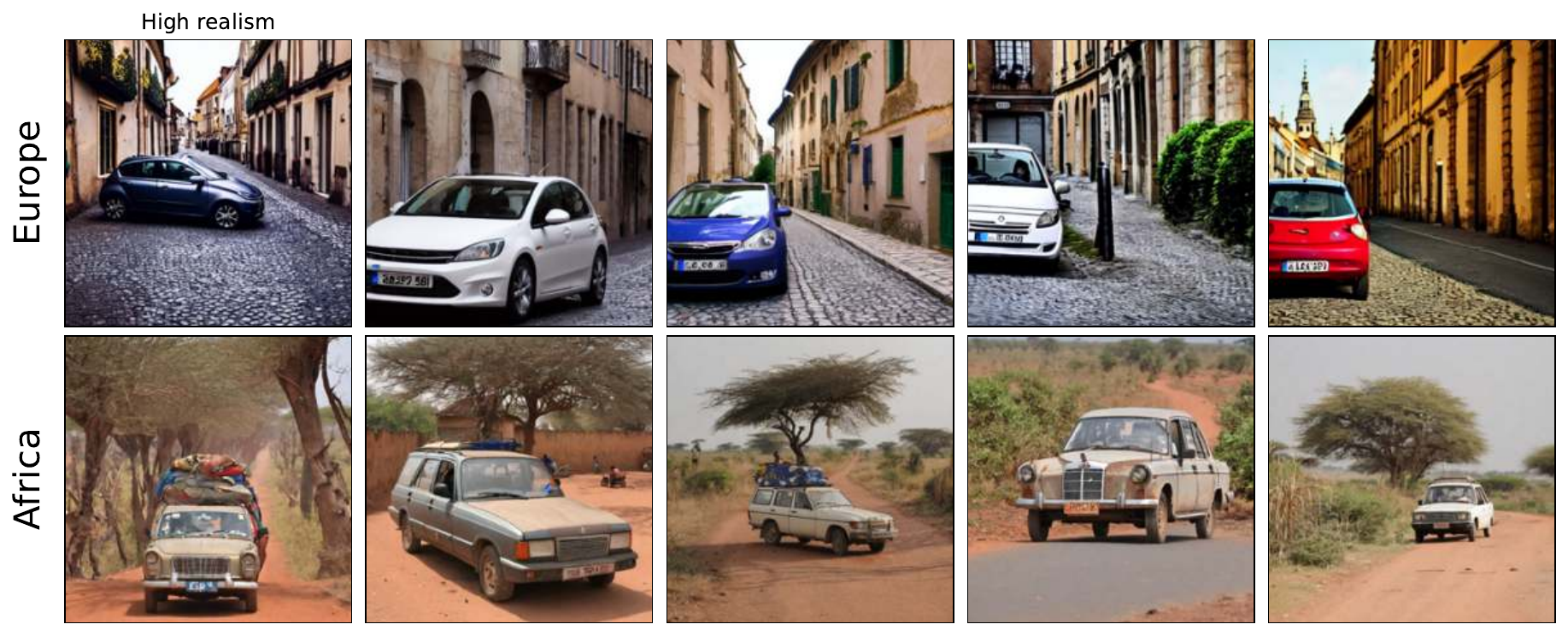}
    \caption{GeoDE qualitative. Left: \texttt{A chair in \{region\}}. Right:  \texttt{A car in \{region\}}}
    \label{fig:quali_geode_pareto}
\end{figure*}

We extend the use of \cdq Pareto fronts to characterize potential geographic disparities of state-of-the-art \cigenmodels. In particular, we follow~\citet{hall2024dig} and investigate geographic disparities of T2I models using the GeoDE dataset~\citep{ramaswamy2024geode}.\looseness-1

\noindent\textbf{\Cd.} \cref{fig:pareto_geode} (left) depicts the region-wise \cd Pareto fronts. We observe that Europe, the Americas, and Southeast Asia exhibit the best Pareto fronts, with consistently higher diversity and consistency than Africa and West Asia. As previously noted, improving diversity (computed as marginal or conditional) comes at the expense of consistency. When considering marginal metrics (top), we observe that Europe and the Americas present the best Pareto fronts. Remarkably, \sdone appears in all region-wise Pareto fronts, whereas \sdtwo appears remarkably less frequently, and does not appear at all in the Pareto front of Europe. This is in line with prior works that demonstrate that recent advancements on standard benchmarks may have come at the cost of reduced real world geographic representations \citep{hall2024dig}. However, we positively discover that disparity \textit{reduction} occurs via \sdxl which appears in the Pareto fronts of Africa, West Asia and South East Asia, bringing the results of Africa closer to those of Europe or the Americas. Yet, \sdxlt only appears in the Pareto fronts of some regions, and presents the highest consistency. We observe that the improvements achieved by \sdxl for Africa are notably reduced when distilling the model into \sdxlt. When considering conditional metrics (bottom), we see that all T2I models appear in the Pareto fronts. Once again, \sdone shows the highest diversity and \sdxlt the highest consistency. As in the previous case, \sdxl only appears in the Pareto fronts of West Asia, Africa, and South East Asia, and bridges the consistency and diversity performance gap between Africa and both Europe and the Americas. Yet, the improvements observed in \sdxl for Africa disappear when considering \sdxlt.\looseness-1

\noindent\textbf{\Qd.} \cref{fig:pareto_geode} (middle) depicts the region-wise \qd Pareto fronts. In the top panel (precision vs. recall), we observe that, similarly to \coco (\cref{fig:paretoT2I}), realism and diversity performance of T2I models present a clear tradeoff. Focusing on the regions, we see that the Pareto fronts of West Asia and Africa are visibly worse than the others. In terms of models, \sdone is the model that generally dominates the Pareto fronts of all regions. Moving to conditional metrics (bottom), we notice similar trends. However, \sdxl appears in the highest realism part of the Pareto front of Africa, and \sdxlt appears in the highest realism part of the Pareto fronts of Europe and Southeast Asia. By looking at the inter-region disparities along different areas of the Pareto fronts, we notice a gradual increase of the inter-region variance when moving from high diversity  (low realism) to high realism (low diversity). This result suggest that maximizing realism might exacerbate stereotypes -- as suggested by the lower diversity -- and increase geographical disparities -- as suggested by the increased variance across region-wise Pareto fronts. We provide a visual validation of this phenomenon in \cref{fig:quali_geode_pareto} (See \cref{suppl:quali_geode_pareto0,suppl:quali_geode_pareto1} in \cref{suppl:results} for more examples). \todo{comment a bit more on the figure here} \looseness-1

\noindent\textbf{\Cq.} \cref{fig:pareto_geode} (right) depics the region-wise \cq Pareto fronts. As shown in the figure, consistency and realism correlate as previously noticed on \coco. The region-wise stratification shows that West Asia and Africa are again the regions with the worst Pareto fronts. The regions that exhibit the best Pareto fronts are East Asia, Southeast Asia, and Europe. Focusing on the top plot (marginal metrics), the Pareto fronts of all regions except the Americas contain \sdone and \sdxlt. Note that \sdone consistently stands out in terms of realism, whereas \sdxlt shines in consistency. \sdtwo and \sdxl are only present in the Pareto of the Americas and Africa, respectively. In the bottom plot (conditional metrics), the situation is very similar, but we notice that for Europe and Southeast Asia the Pareto is only composed by \sdxlt.

\begin{tcolorbox}[my_style]
\subsubsection*{Key insights}
\begin{itemize}[leftmargin=.3em]
    \item Improving generation diversity comes at the expense of consistency for all regions considered. Realism and diversity also present a clear tradeoff for all regions, whereas realism and consistency appear correlated.
    \item Interestingly, the oldest model, \sdone dominates the most recent ones, and consistently appears in the Pareto fronts of all regions, when considering any pair-wise objective. However, \sdxl reduces the disparities between Africa and Europe or the Americas in terms of diversity and consistency, as we move towards the high consistency part of the Pareto fronts.
    \item Advances in T2I models reduce region-wise disparities in terms of consistency and increase the disparities in terms of realism, while sacrificing diversity across all regions.
\end{itemize}
\end{tcolorbox}

\subsection{The impact of knobs on \cdq}
\label{sec:knobParetos}

Finally, in this section, we study the effect of different knobs that control consistency, diversity and realism of \cigenmodels.  In the interest of space, we focus on the conditional metrics, and perform the analysis on \coco. 

\begin{figure*}[ht]
    \centering
    \includegraphics[height=.55cm]{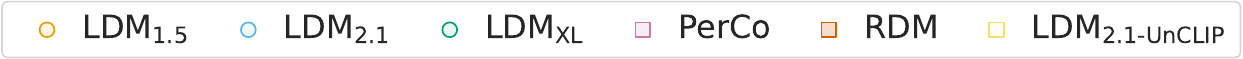}
    \includegraphics[height=.55cm]{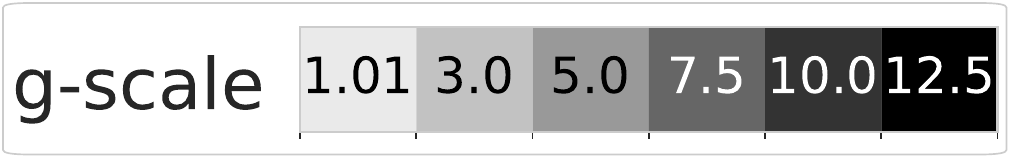}
    \includegraphics[width=\textwidth]{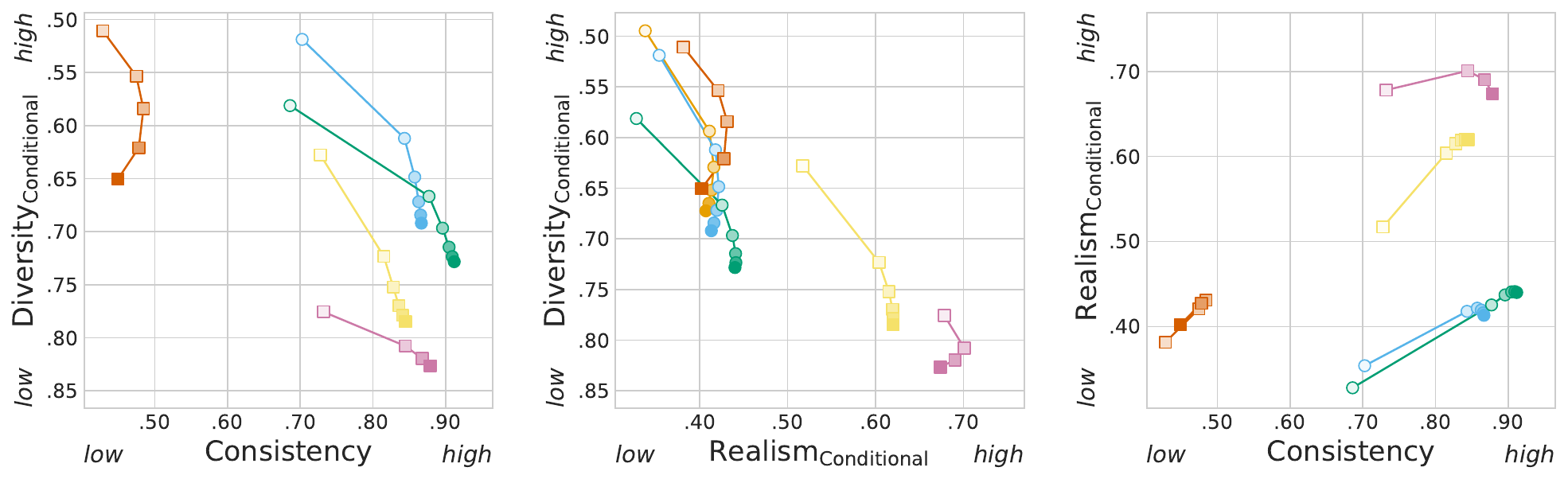}
    \caption{Ablation on guidance scale. To help readability, we report only a subset of the points presented in \cref{fig:paretoT2I,fig:paretoI-T2I}, selecting runs with default values for other knobs. }
    \label{fig:gscale}
    \vspace{-1em}
\end{figure*}
\begin{figure*}[ht]
    \centering
    \includegraphics[height=.55cm]{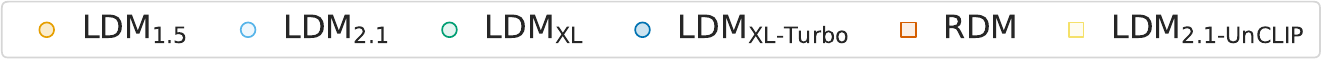}
    \includegraphics[height=.55cm]{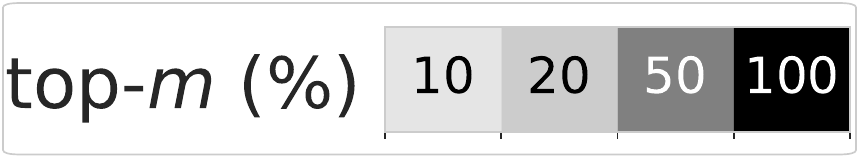}
    \includegraphics[width=\textwidth]{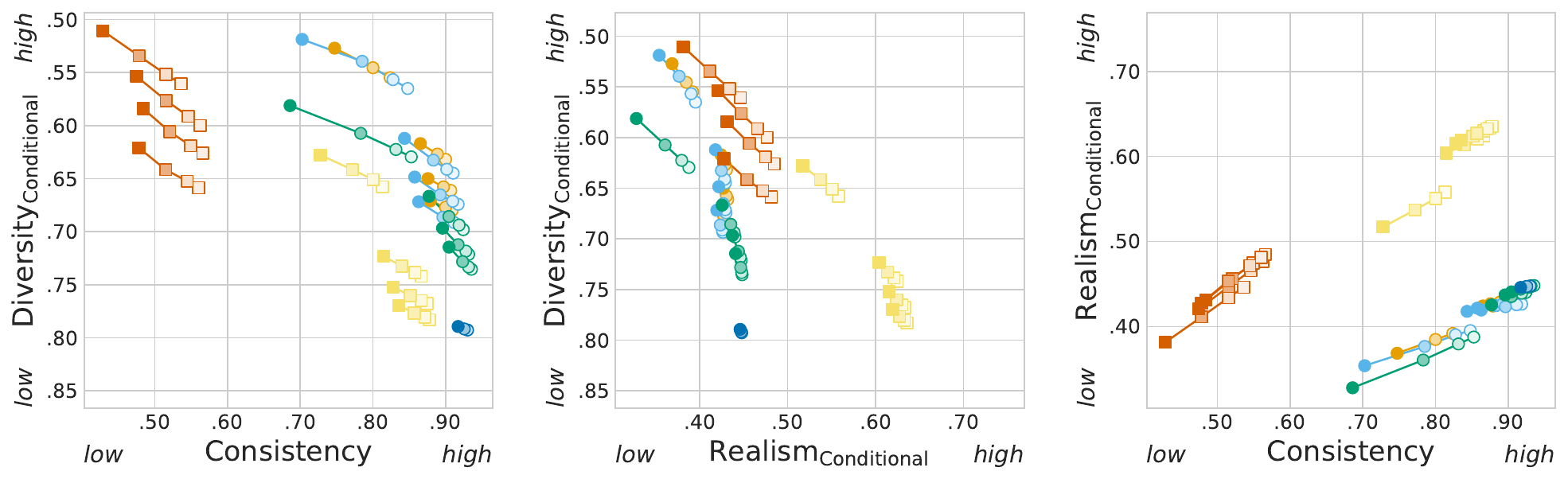}
    \caption{Ablation on \ktopk.  }
    \label{fig:ktopk}
\end{figure*}

\noindent\textbf{Guidance scale.}
\cref{fig:gscale} depicts the effect of guidance scale on \cd (left panel), \qd (middle panel), and \cq (right panel) objectives. By looking at the \cd plot, we observe that increasing the guidance scale leads to improved consistency at the expense of the diversity in most cases~\footnote{\todo{why RDM no}}, with \sdxl showing the highest relative improvements. Moreover, for all models we notice that the initial increase in the guidance scale -- from 1.01 to 3.0 -- leads to the biggest consistency improvements.  
By looking at the \qd plot, we note that the increase in the guidance scale often leads to increase in realism at the expense of diversity, with \unclip and \perco benefiting the most and the least from this knob, respectively. Moreover, we note that, in most cases, increasing the guidance scales beyond $~$$7.5$ no longer results in realism improvements. Finally, the \cq plot reveals that by increasing the guidance scale the models generally improve both the consistency and realism. However, too large values of guidance may lead to decreasing the image realism; this happens for all models except of \unclip and \sdxl.\looseness-1 %

\noindent\textbf{Post-hoc filtering.}
\cref{fig:ktopk} depicts the effect of applying \ktopk. 
In the \cd plot (left), we observe that \ktopk (based on \clipscore) leads to improvements in consistency for all models -- the lower the value of $m$, the higher the consistency. Unsurprisingly, the models that initially have high consistency scores do not gain as much when leveraging \ktopk as the models that start with low consistency scores. Moreover, we observe that the post-hoc filtering consistently leads to a diversity decrease. However, this decrease is less pronounced for the \ktopk than for the guidance knob, as is the case for the consistency increase (\cf \cref{fig:gscale}). 
The diversity-realism plot (middle) shows that post-hoc image filtering leads to an increase in the realism at the expense of diversity. By looking at the realism-consistency plot (right), we note that the post-hoc filtering is an effective way to increase both image realism and consistency, with the latter one improving faster.\looseness-1 %

\begin{figure*}[ht]
    \centering
    \includegraphics[height=.55cm]{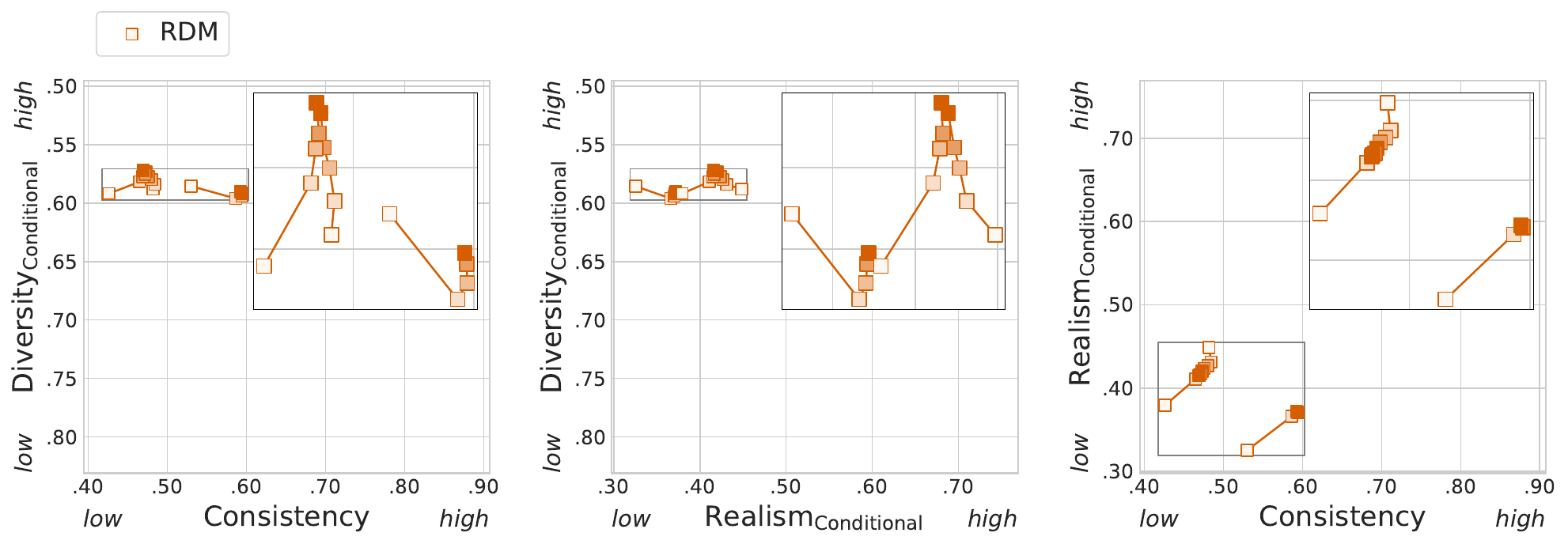}
    \includegraphics[height=.55cm]{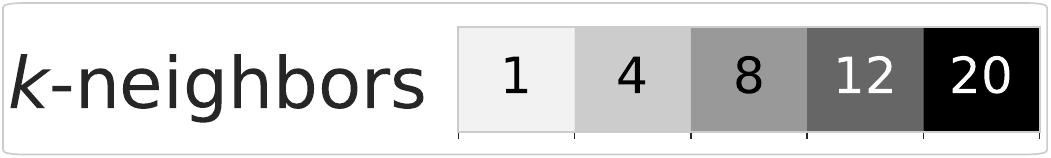}
    \includegraphics[width=\textwidth]{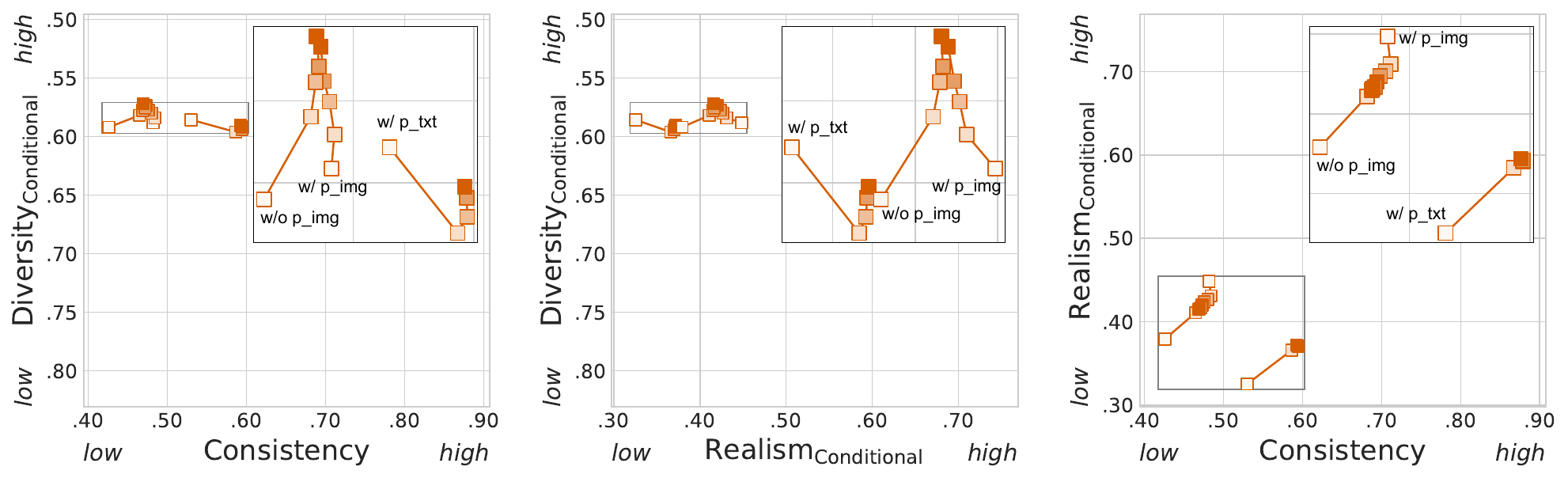}
    \caption{The effect of  the neighborhood size on  diversity, consistency and realism metrics. To improve readability we report a zoomed-in view in the top right of each plot.}
    \label{fig:knn}
    \vspace{-1em}
\end{figure*}

\begin{figure*}[ht]
    \centering
    \includegraphics[height=.55cm]{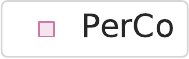}
    \includegraphics[height=.54cm]{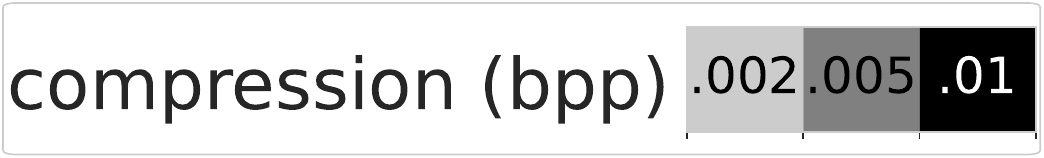}
    \includegraphics[width=\textwidth]{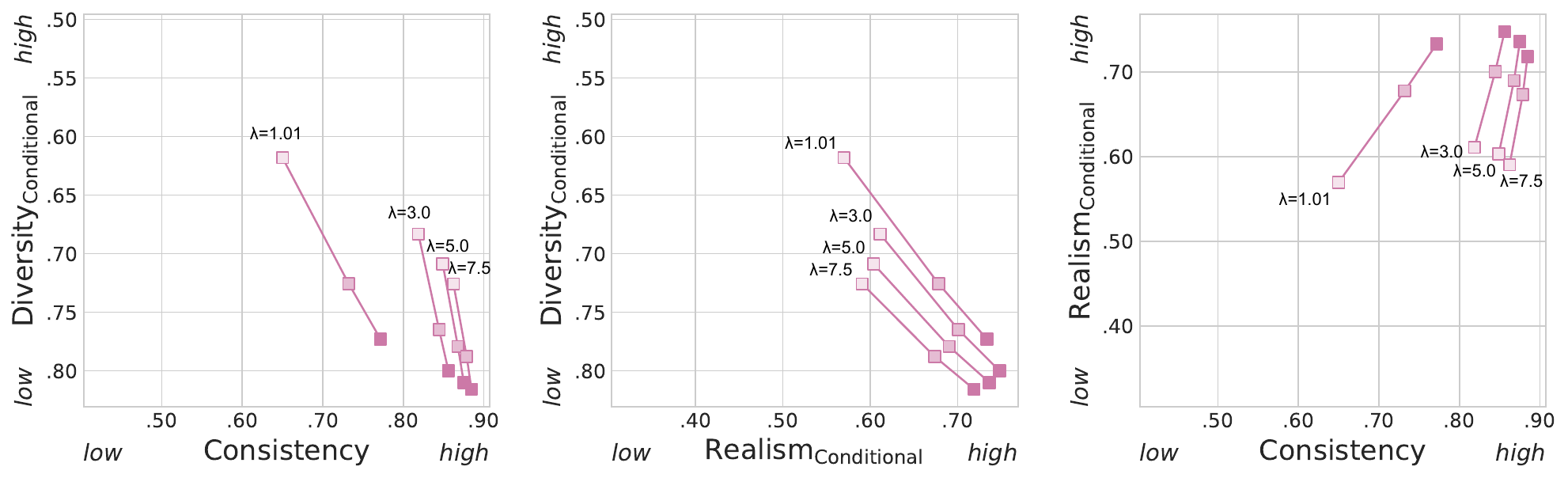}
    \caption{The effect of the compression rate on  diversity, consistency and realism metrics.}
    \label{fig:bpp}
    \vspace{-1em}
\end{figure*}

\noindent\textbf{Retrieval augmentation neighborhood size.}
The amount of neighbors used in retrieval augmentation may impact consistency, diversity, realism based on the semantic of the neighbors. In \cref{fig:knn}, we study the impact of the neighborhood size $k$ for \rdm. 
We notice that, in absolute terms, the impact of $k$ is minor in all the pairs of metrics considered, suggesting that this knob is not as effective as the previous ones. \todo{Revisit claims below, when knowing the axes values on the zoomed in plot} In the \cd plot (left), we observe that increasing $k$ from $4$ to $20$ leads to a small but consistent increase in diversity, while maintaining consistency. However, when increasing $k$ from $1$ to $4$, we generally see a small improvement in consistency. This result is expected as by increasing the neighborhood size we might include more diverse neighbors, and as long as those neighbors are semantically similar to the query image, they will not affect the consistency of the generation. In the \qd plot (middle), we observe similar trends: increasing $k$ from 4 to 20 results in small diversity improvements with little to no effect on realism, while increasing $k$ from 1 to 4 results in small realism improvements. Interestingly, \rdm prompted with text achieves lower realism than the others models. Moreover, increasing $k$ when the query image is present together with the neighbors, slightly harms the realism. %
Finally, in the \cq plot (right), we note a positive correlation between the two metrics when text query or no query is used.\looseness-1 %

\noindent\textbf{Compression rate.}
The reconstructions produced by an image compression model are highly dependent on the selected compression rate,  measured  in terms of bit-per-pixel (bbp) of the compressed image, where high compression rate means low bpp. 
In \cref{fig:bpp} we assess \perco with different bitrates and  at different guidance scales. 
By looking at the left panel, we observe that decreasing the bitrate leads to notable increases in conditional diversity, which is inline with qualitative observations made by \citet{careil2024towards}. 
Moreover, these diversity increases only marginally reduce consistency, especially for guidance scales $> 3$, suggesting that even at high compression rates, the reconstructed images maintain their semantics. By contrast, in \qd (middle), higher compression leads to a pronounced loss in realism, suggesting that the reconstructed images do not necessarily capture all the details from the original images.
Finally, the results presented for \cq (right) suggest, once again, that consistency and realism are correlated. 

\begin{tcolorbox}[my_style]
\begin{itemize}[leftmargin=.3em]
    \item Guidance scale trades diversity for consistency and realism. Consistency and realism improve with higher guidance scale, but realism improvements saturate earlier than consistency improvements.
    \item Post-hoc filtering improves consistency and realism at the expense of diversity. Although both consistency and realism improve with this knob, consistency increases at a faster pace. Overall, post-hoc filtering appears less effective than guidance scale.
    \item The effect of retrieval augmentation on \cdq appears minor, questioning the knobs efficacy to control the multi-objective. 
    \item Compression rate affects image realism and diversity, but has little effect on consistency, as compression models tend to maintain the image semantics. 
\end{itemize}
\end{tcolorbox}
\section{Conclusions}
\label{sec:conclusion}
We proposed \cdq Pareto fronts as a comprehensive framework to evaluate \cigenmodels and their potential as visual world models. Using this framework, we have been able to compare several existing models on the consistency-diversity-realism axes, which allowed us to provide insights on which model is preferable over another based on the objective at hand. Our results highlighted the presence of tradeoffs among the consistency-diversity and realism-diversity axes in all the studied models. In particular, we discovered an interesting trend in the historical/temporal evolution of image generative modes, with earlier models (\eg, \sdone and \sdtwo) achieving higher diversity and more balanced tradeoffs than latest models (\eg \sdxl), which instead trade diversity to favour consistency and realism. All in all, our analysis suggested that there is no best model and the choice of model should be determined by the downstream application.
We hope that Pareto fronts will become a new standard for evaluating the potential of \cigenmodels as world models.

\noindent\textbf{Limitations.} Our analysis only considers open models as evaluating closed models is very expensive or sometimes not possible. It would be interesting placing the dots of closed state-of-the-art models within the multi-objective pareto front. Moreover, it would be interesting to extend the analysis to ablate further knobs. For example, we have not included the knob of structured conditioning, like layouts, sketches or other form of control typically used to increase consistency. Another aspect that our analysis does not ablate is the effect of different data distribution on the consistency-diversity-realism pareto fronts --this aspect is currenty very hard to study  due to the closed data filtering recipes of most models. Furthermore, for certain evaluated knobs like the retrieval augmented generation, the analysis could be deepen by considering for example the effect of different retrieval databases or stronger/more recent models than \rdm ---unfortunately, there is a scarcity of open models using RAG. Finally, our work suggests future research to understand whether the observed tradeoffs are fundamental, or could be overcome by future generations of better generative models.

\bibliography{main}
\bibliographystyle{tmlr-template/tmlr}

\clearpage
\newpage
\beginappendix
\section{Implementation details}
\label{suppl:details}

\cref{suppl:tab:knob_values} reports the exact knob values ablated for each model.

\begin{table}[ht]
    \centering
    \caption{Knob values ablated per model.}
    \begin{tabular}{l|l}
    \toprule
         Knob &  values \\
    \midrule
         \multirow{2}{*}{\kscale} & All LDM models: $[1.01, 3.0, 5.0, 7.5, 10.0, 12.5]$; \\
             & \rdm: $[1.01, 1.5, 2.0, 3.0, 5.0]$ \\
             & \perco: $[1.01, 3.0, 5.0, 7.5]$ \\
    \midrule
         \ktopk & All but \perco: $[10, 20, 50, 100]\%$ \\
    \midrule
         \kneighbor & \rdm: $[1, 4, 8, 12, 20]$ \\
    \midrule
        \kcomp & \perco: $[0.01, 0.005, 0.002]$bpp \\
    \bottomrule
    \end{tabular}
    \label{suppl:tab:knob_values}
\end{table}

\section{Additional results}
\label{suppl:results}

\todo{correlation plots, and plots for geode}

\subsection{Additional T2I results on \coco}
\begin{figure*}[ht]
    \centering
    \begin{subfigure}[b]{\textwidth}
    \includegraphics[width=\textwidth]{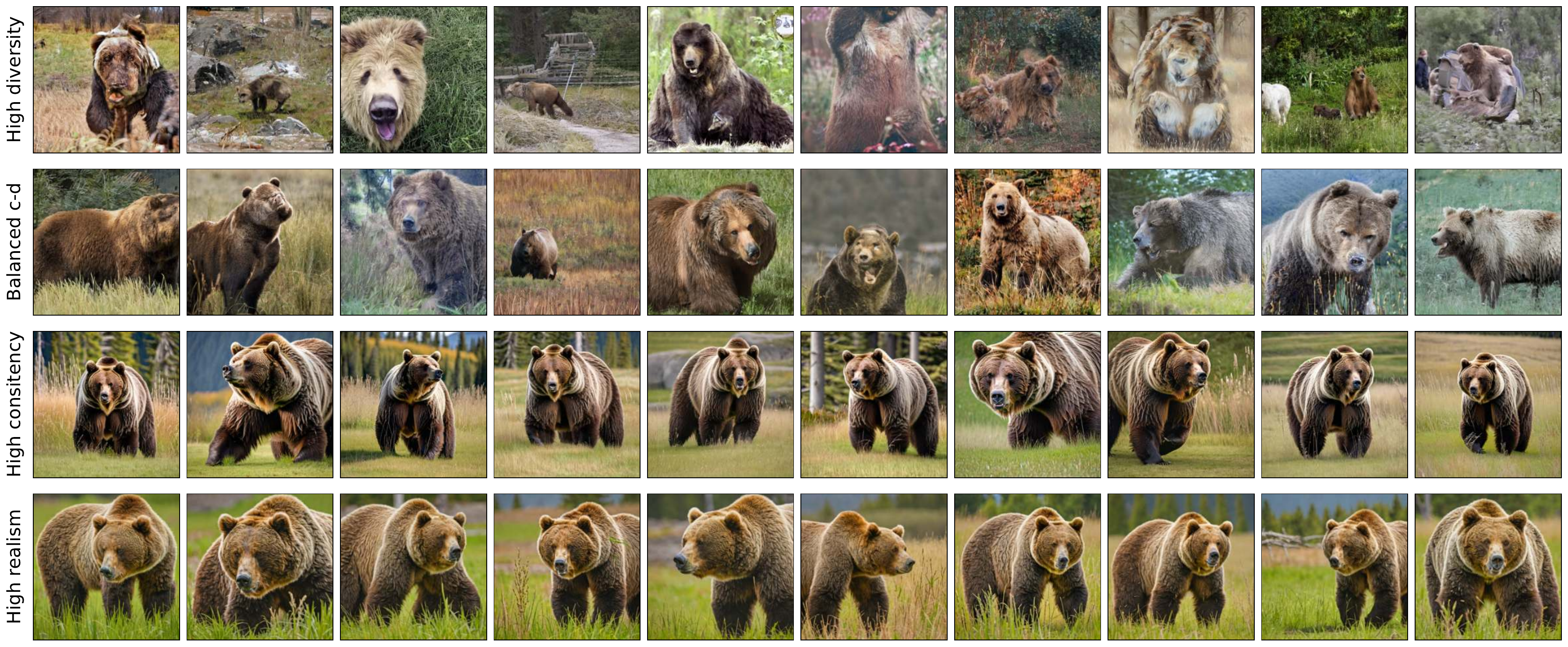}
    \caption{$\rvp:$ \texttt{A big burly grizzly bear is show with grass in the background.}}
    \end{subfigure}
    \begin{subfigure}[b]{\textwidth}
    \includegraphics[width=\textwidth]{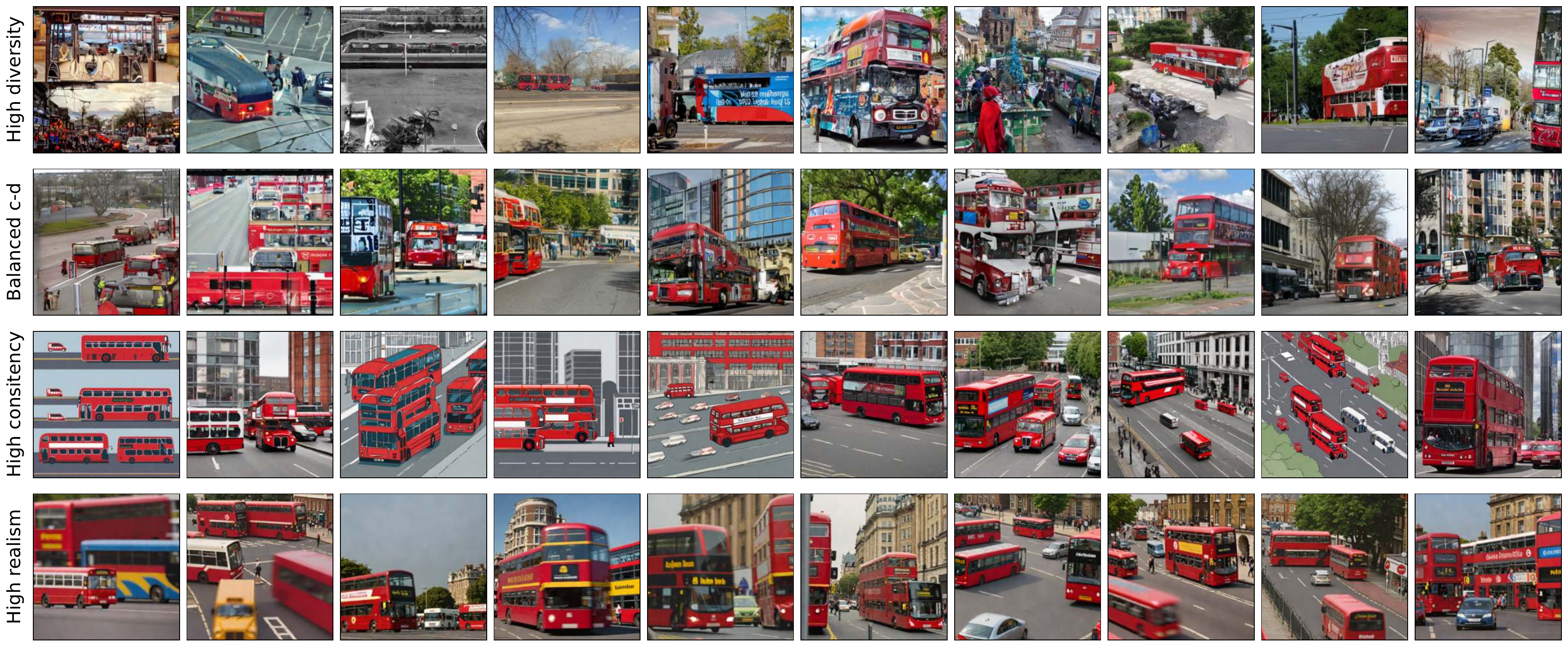}
    \caption{$\rvp:$ \texttt{The red, double decker bus is driving past other buses. }}
    \end{subfigure}
    \begin{subfigure}[b]{\textwidth}
    \includegraphics[width=\textwidth]{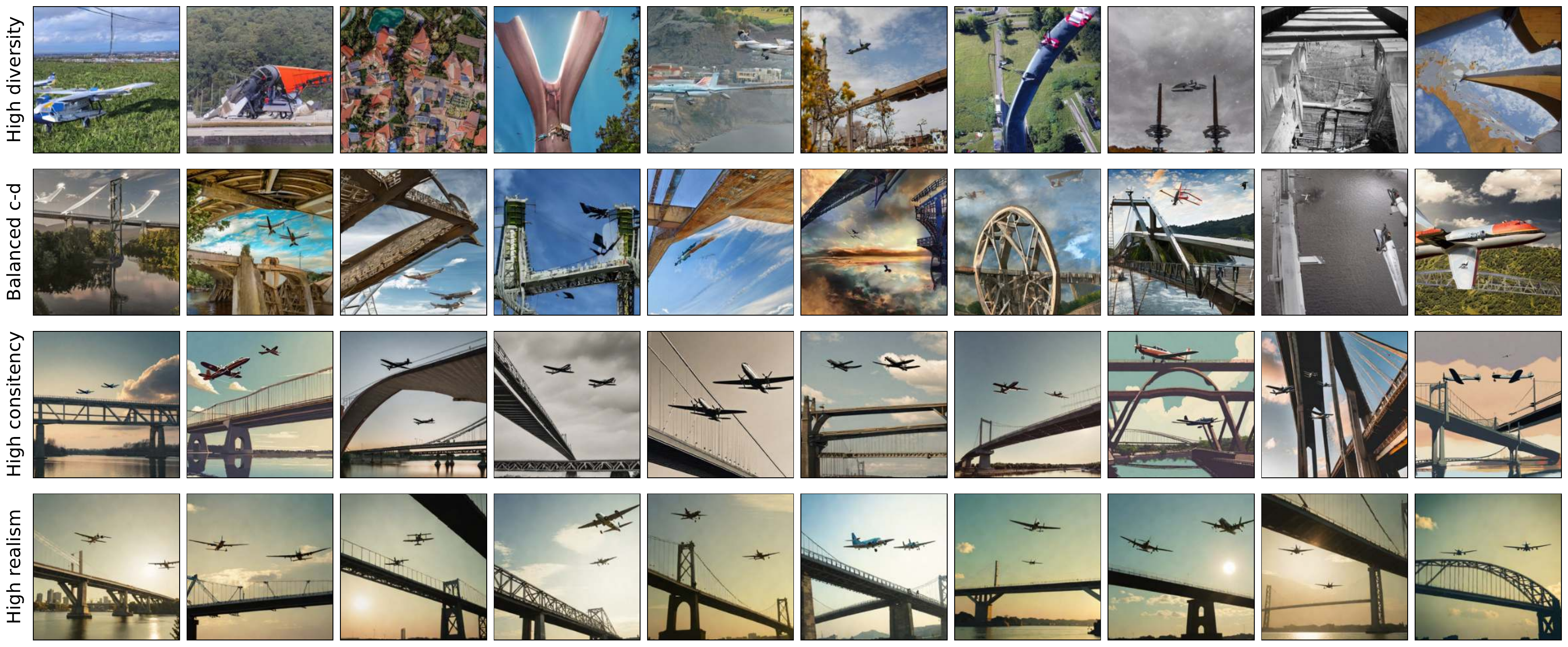}
    \caption{$\rvp:$ \texttt{Two planes flying in the sky over a bridge.
}}
    \end{subfigure}
    \caption{``High diversity'': \sdone; ``Balanced c-d'': \sdtwo; ``High consistency'': \sdxl, ``High consistency'': \sdxlt}
    \label{suppl:quali_t2i_pareto1}
\end{figure*}

\begin{figure*}[ht]
    \centering
    \begin{subfigure}[b]{\textwidth}
    \includegraphics[width=\textwidth]{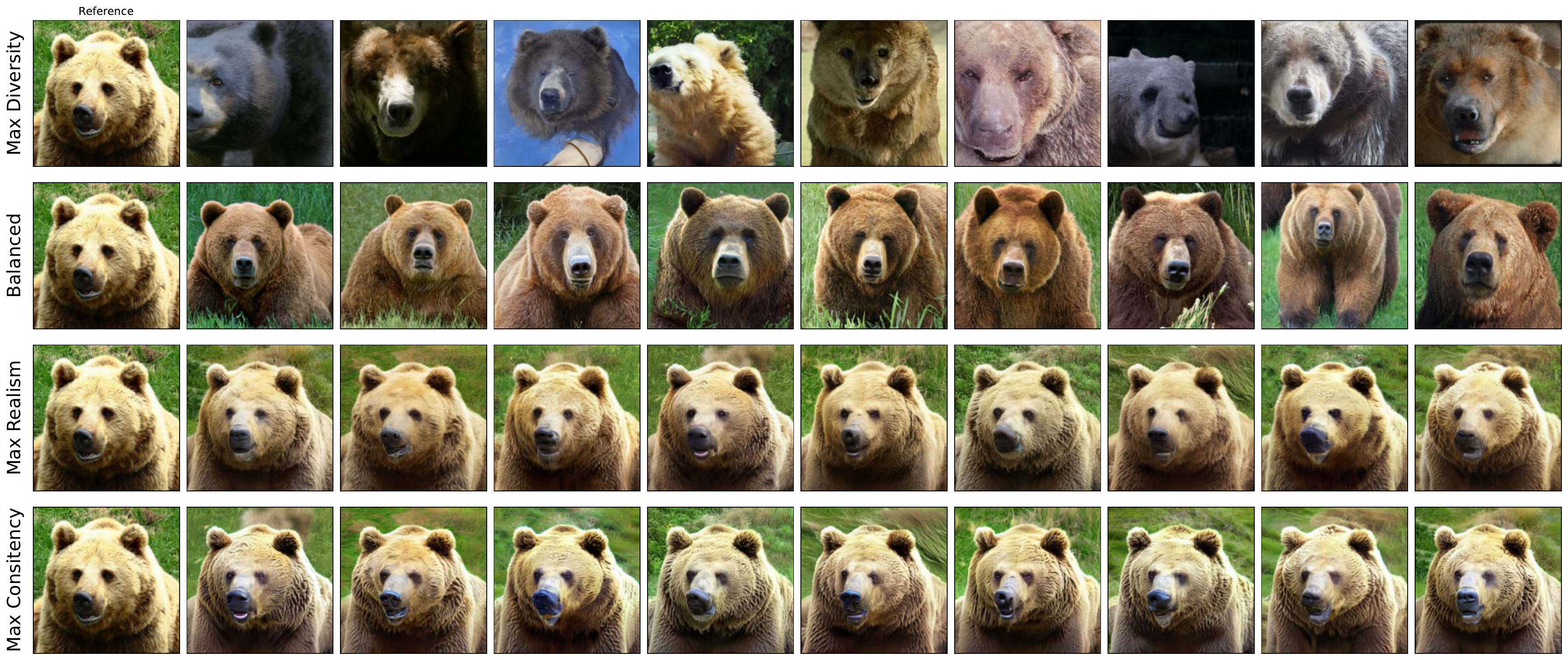}
    \caption{$\rvp:$ \texttt{A big burly grizzly bear is show with grass in the background.}}
    \end{subfigure}

    \begin{subfigure}[b]{\textwidth}
    \includegraphics[width=\textwidth]{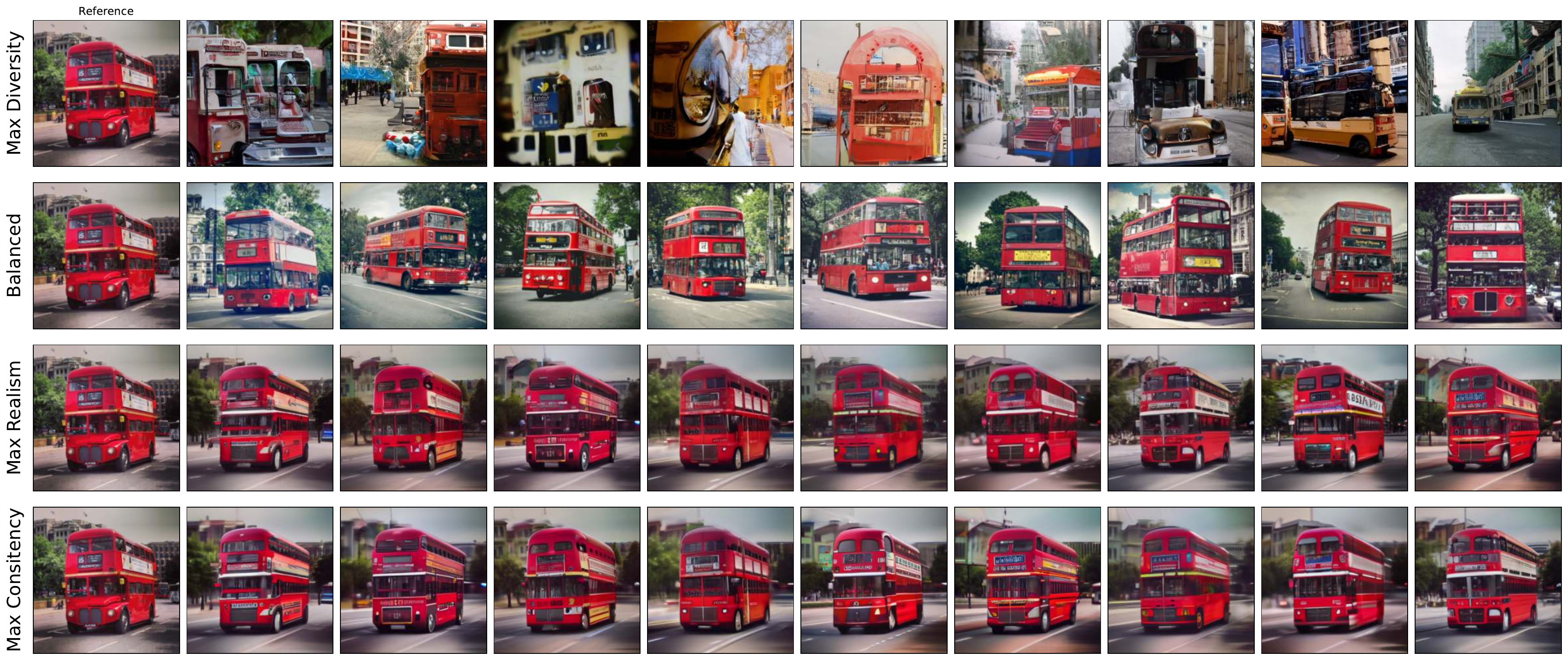}
    \caption{$\rvp:$ \texttt{The red, double decker bus is driving past other buses. }}
    \end{subfigure}
    \begin{subfigure}[b]{\textwidth}
    \includegraphics[width=\textwidth]{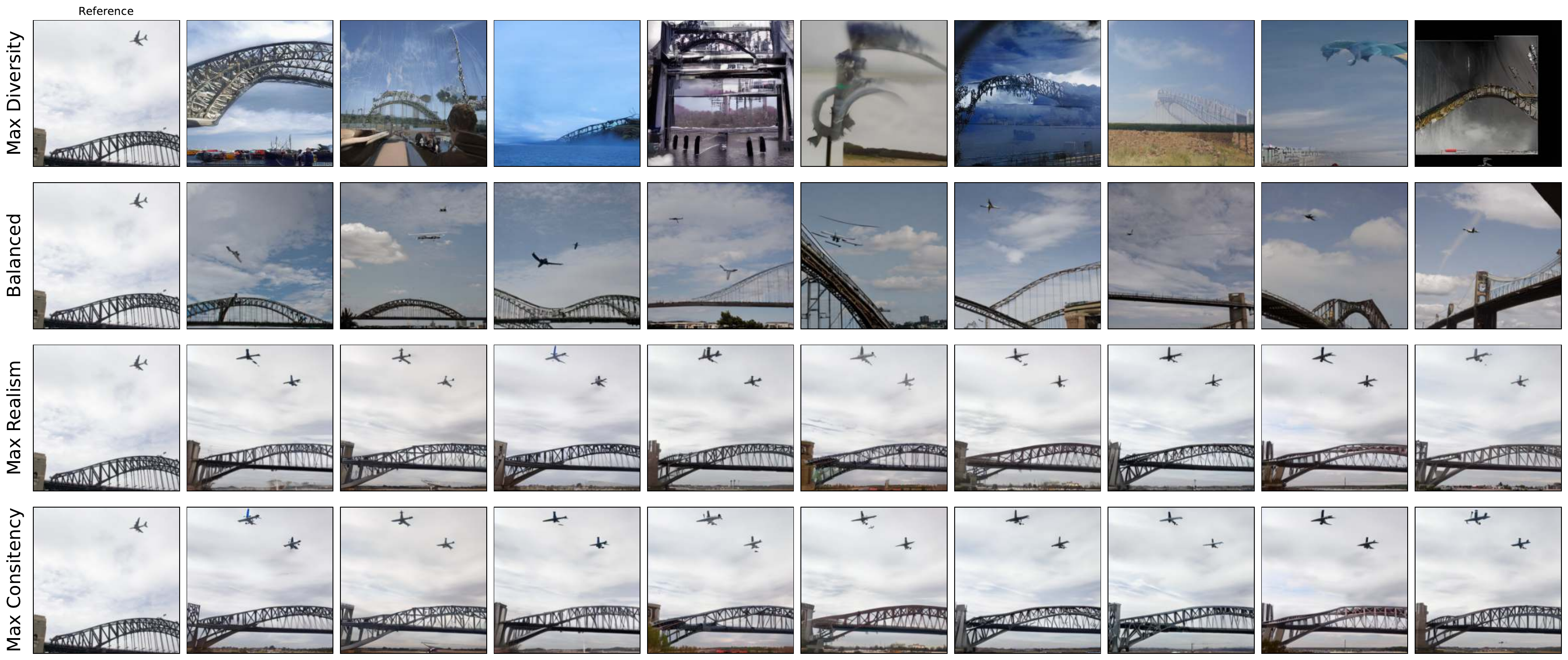}
    \caption{$\rvp:$ \texttt{Two planes flying in the sky over a bridge.}}
    \end{subfigure}

    \caption{``High Diversity'': \rdm; ``Balanced c-d'': \unclip; ``High consistency and realism'': \perco}
    \label{suppl:quali_it2i_pareto1}
\end{figure*}

\begin{figure*}[ht]
    \centering

    \begin{subfigure}[b]{\textwidth}
    \includegraphics[width=\textwidth]{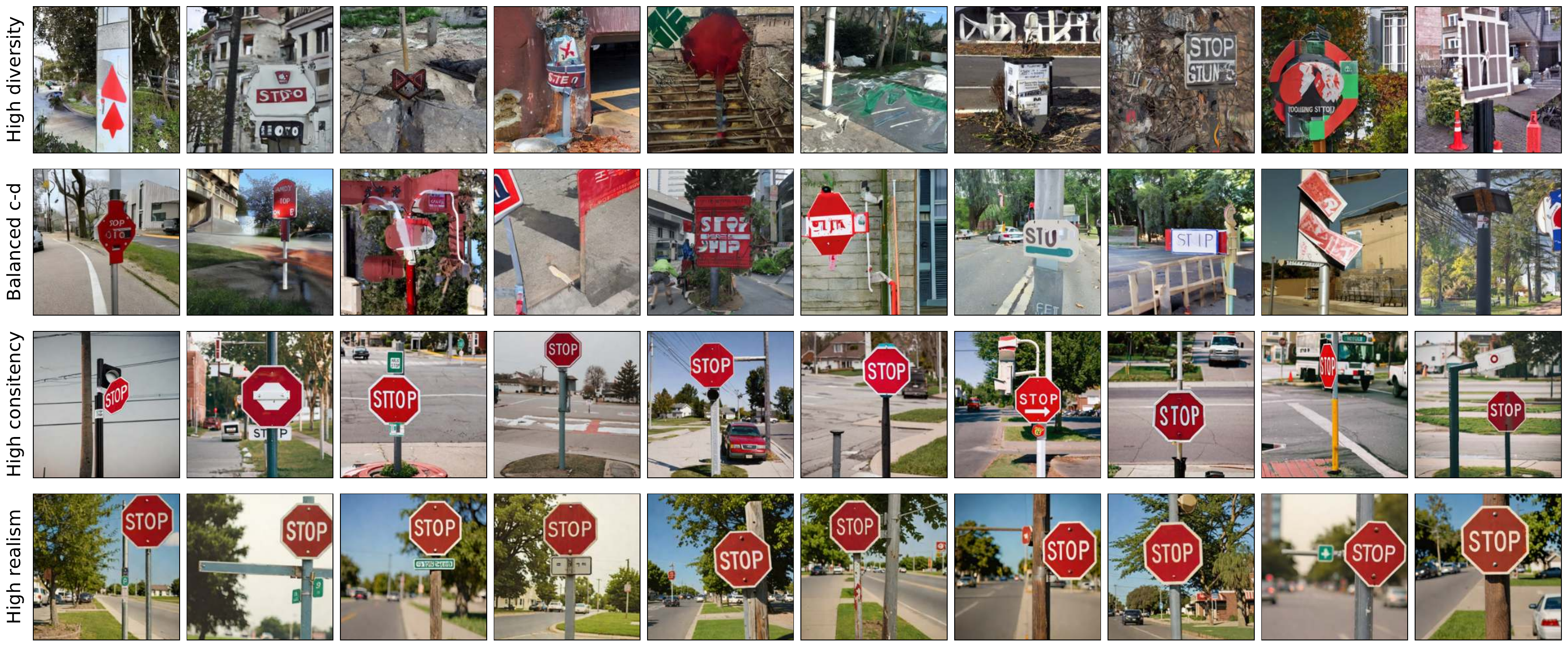}
    \caption{$\rvp:$ \texttt{A stop sign is mounted upside-down on it's post. }}
    \end{subfigure}

    \begin{subfigure}[b]{\textwidth}
    \includegraphics[width=\textwidth]{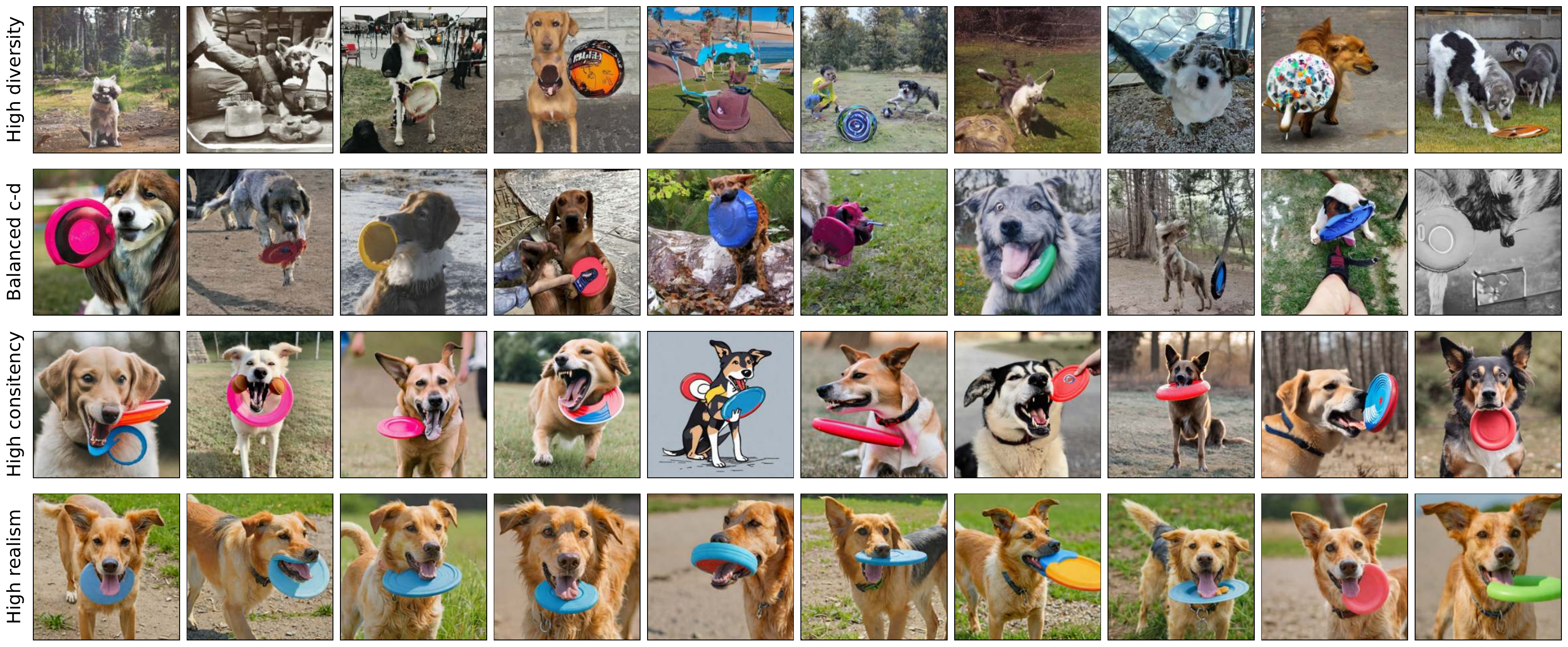}
    \caption{$\rvp:$ \texttt{There is a dog holding a Frisbee in its mouth.}}
    \end{subfigure}
    
    \begin{subfigure}[b]{\textwidth}
    \includegraphics[width=\textwidth]{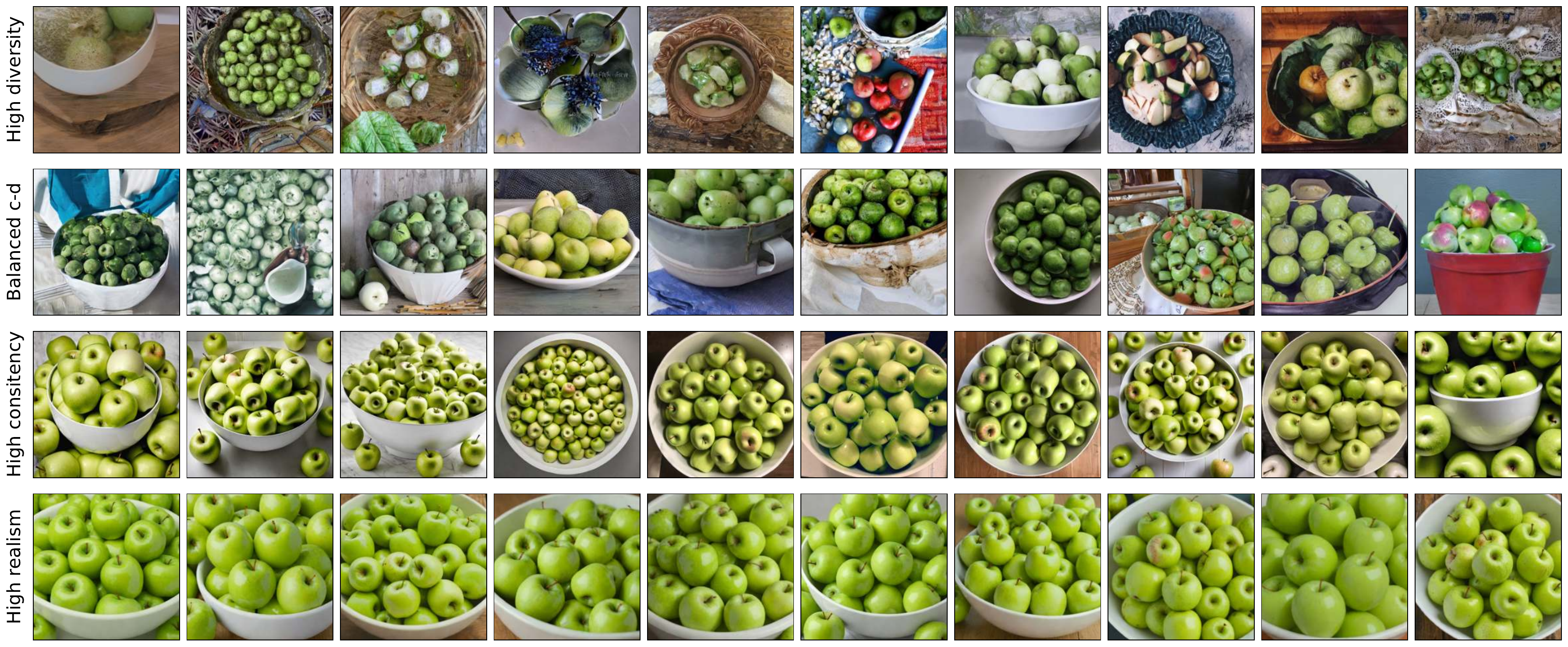}
    \caption{$\rvp:$ \texttt{A large white bowl of many green apples.}}
    \end{subfigure}
    \caption{``High diversity'': \sdone; ``Balanced c-d'': \sdtwo; ``High consistency'': \sdxl, ``High consistency'': \sdxlt}
    \label{suppl:quali_t2i_pareto2}
\end{figure*}

\begin{figure*}[ht]
    \centering
    
    \begin{subfigure}[b]{\textwidth}
    \includegraphics[width=\textwidth]{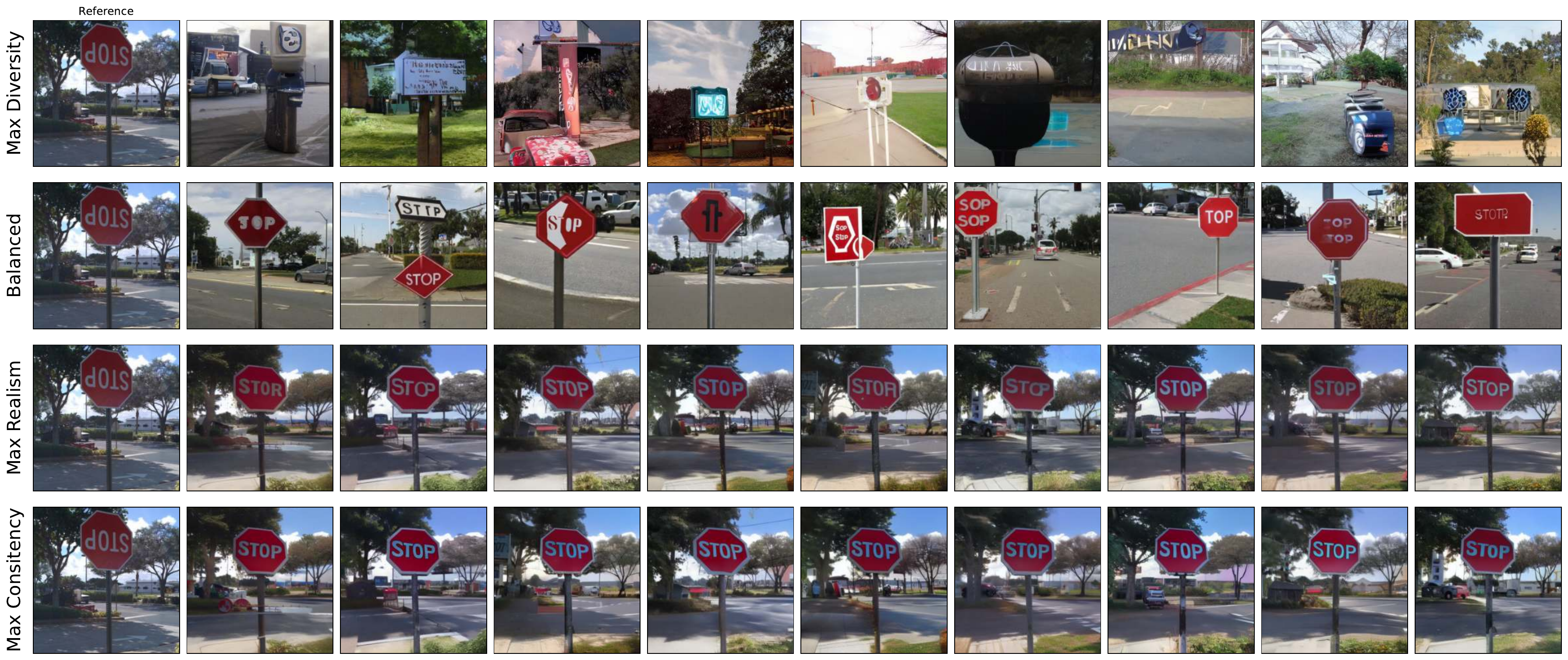}
    \caption{$\rvp:$ \texttt{A stop sign is mounted upside-down on it's post. }}
    \end{subfigure}

    \begin{subfigure}[b]{\textwidth}
    \includegraphics[width=\textwidth]{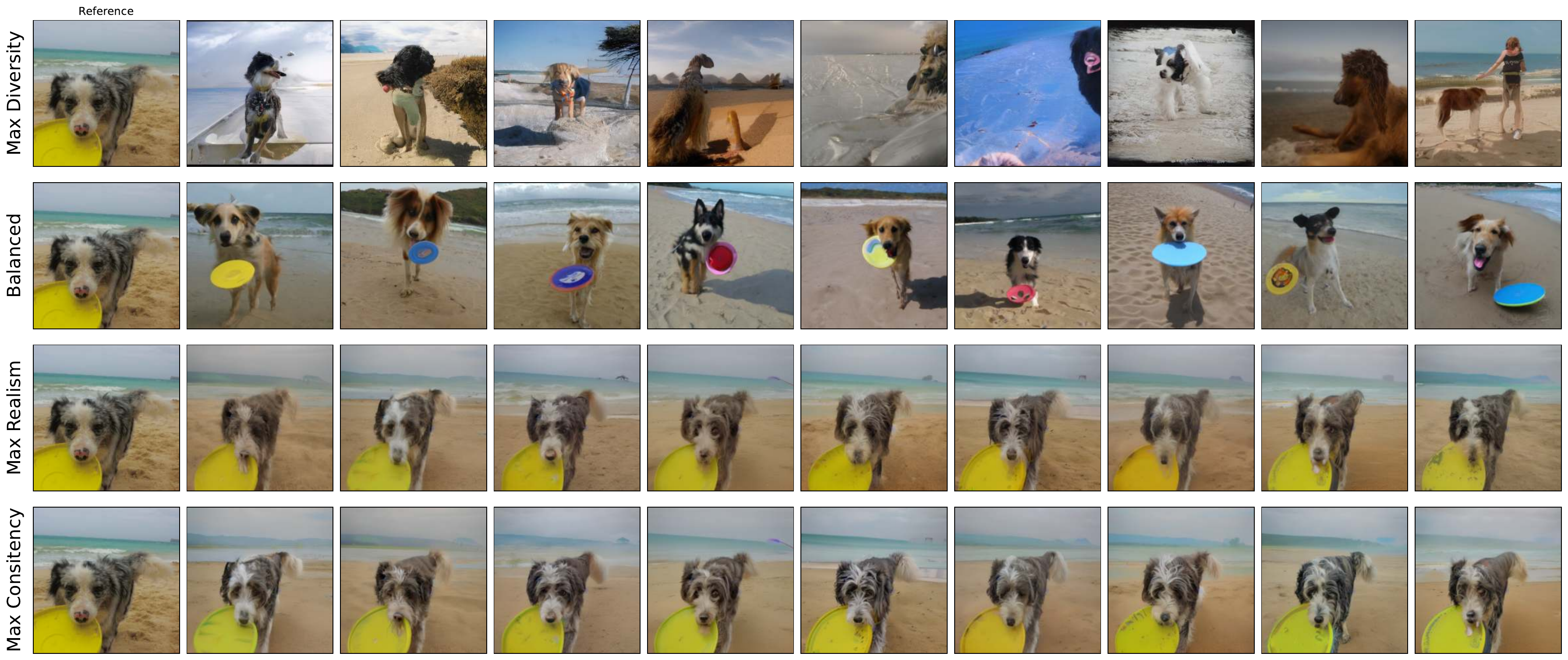}
    \caption{$\rvp:$ \texttt{There is a dog holding a Frisbee in its mouth.}}
    \end{subfigure}
    
    \begin{subfigure}[b]{\textwidth}
    \includegraphics[width=\textwidth]{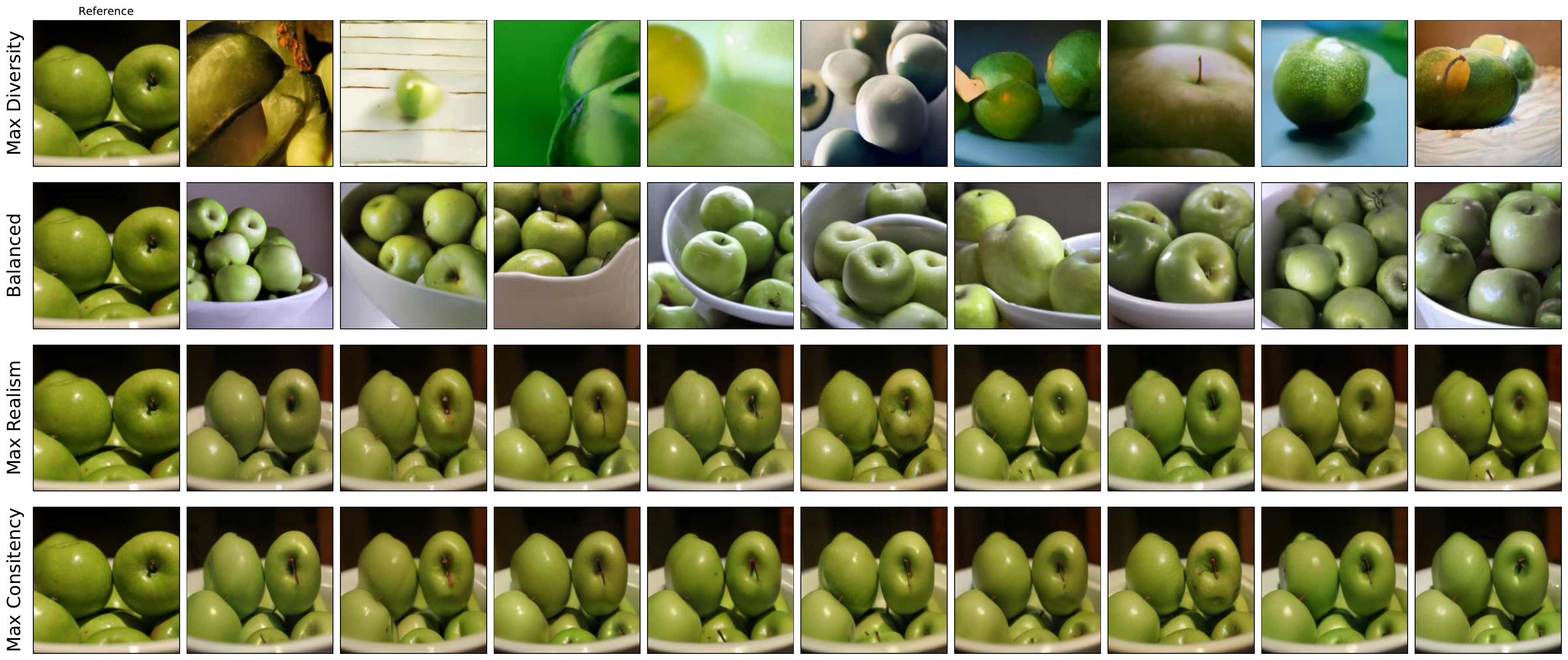}
    \caption{$\rvp:$ \texttt{A large white bowl of many green apples.}}
    \end{subfigure}
    \caption{``High diversity'': \rdm; ``Balanced c-d'': \unclip; ``High consistency and realism'': \perco}
    \label{suppl:quali_it2i_pareto2}
\end{figure*}

\noindent\textbf{Additional qualitative.} \cref{suppl:quali_t2i_pareto1,suppl:quali_it2i_pareto1,suppl:quali_t2i_pareto2,suppl:quali_it2i_pareto2} depict images generated with models present in the Pareto fronts at different locations. Four models are chosen in order to provide exemplars of different areas of the Pareto: one model has the highest diversity, one has balanced consistency-diversity or realism-diversity, one has the highest consistency, and one has the highest realism. The visual comparison of the different models (different rows) validates a noticeable difference among the models in terms of consistency, diversity, and realism. Moreover, we notice that in the case of highest diversity the models tend to generate noisy images, sometimes hardly relatable with the prompt.

\begin{figure}[ht]
    \centering
    \includegraphics[width=\textwidth]{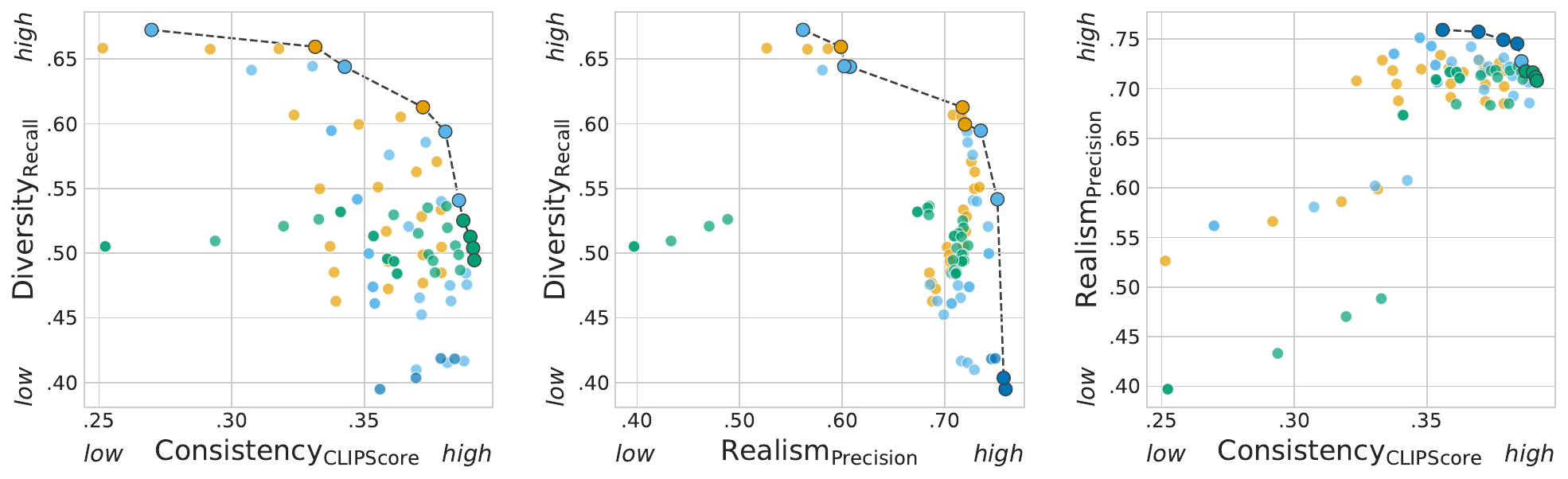}
    \includegraphics[width=\textwidth]{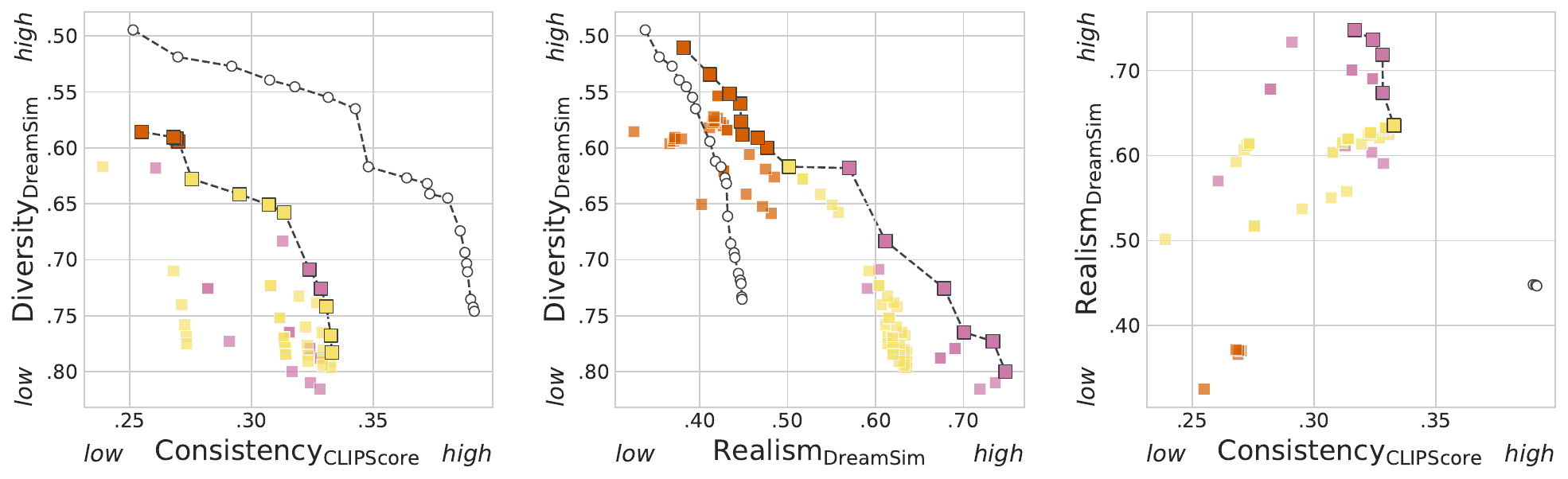}
    \centering
    \includegraphics[width=\textwidth]{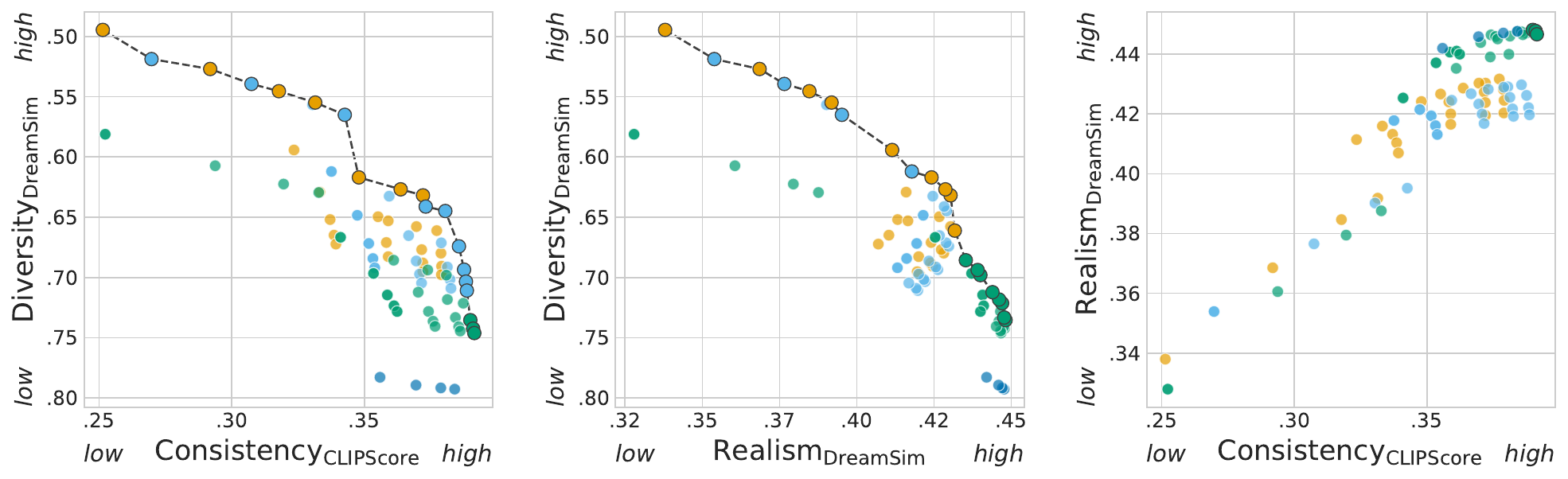}
    \includegraphics[width=\textwidth]{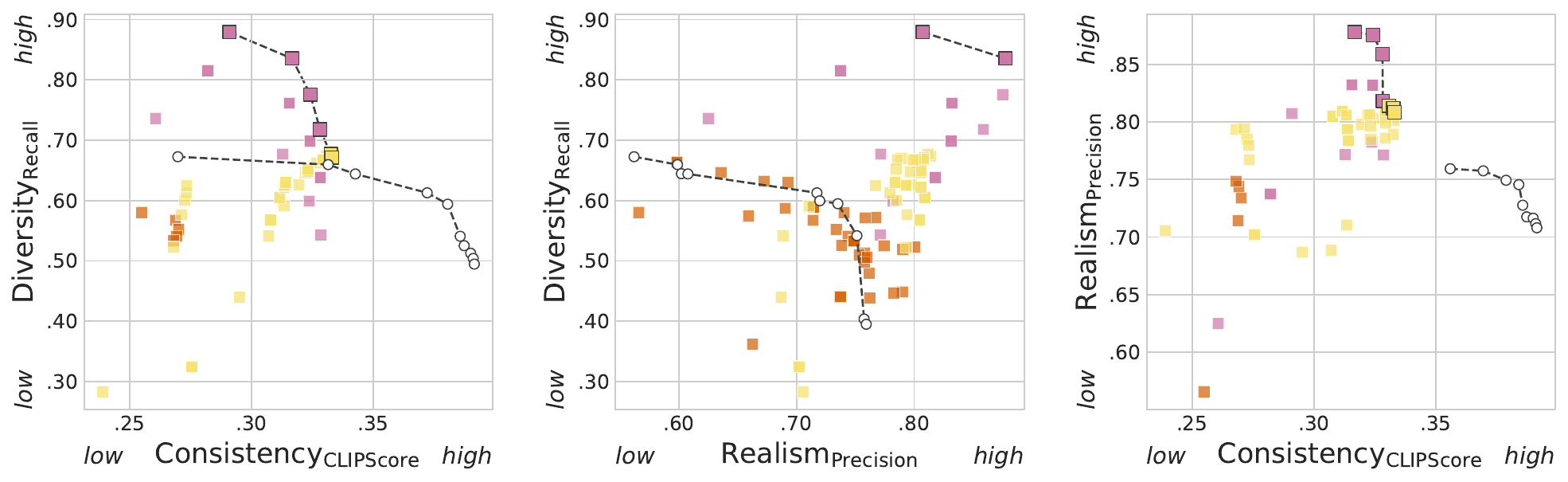}
    \caption{Using \clipscore for consistency.}
    \label{suppl:metrics1}
\end{figure}

\begin{figure}[ht]
    \centering
    \includegraphics[width=\textwidth]{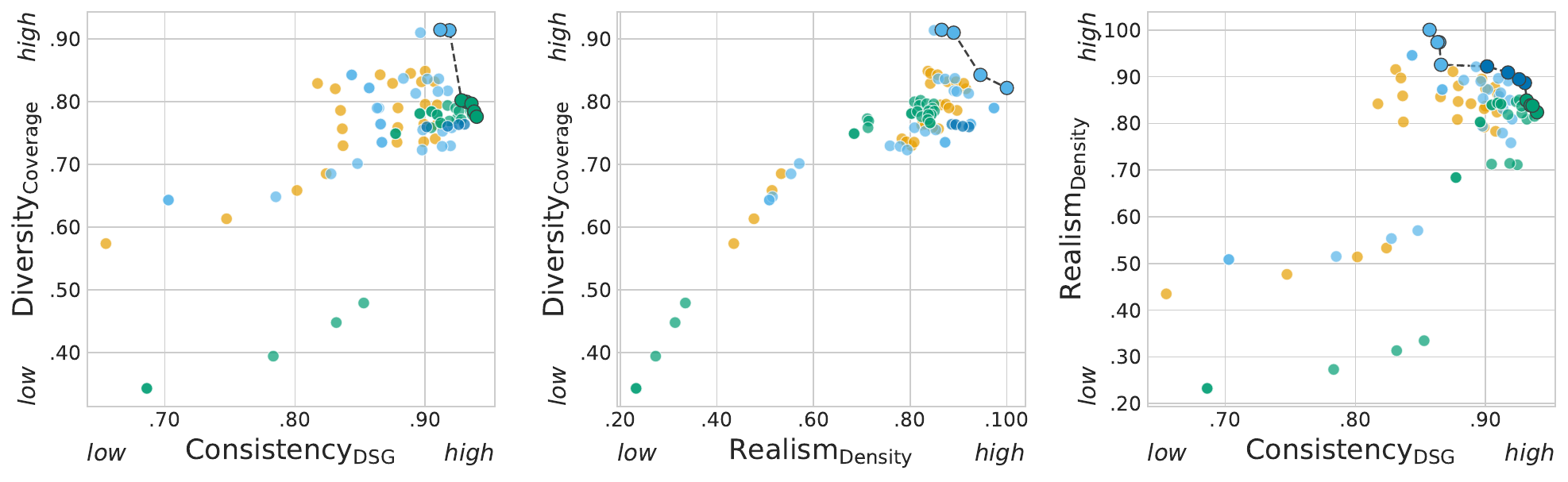}
    \includegraphics[width=\textwidth]{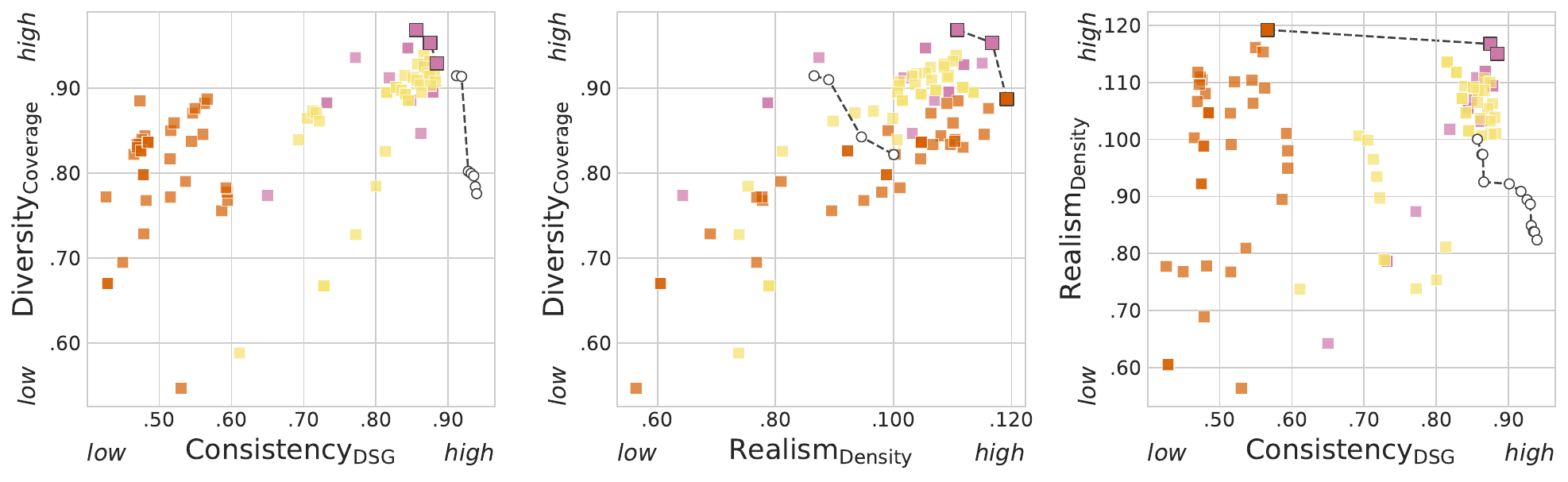}
    \caption{Using density and coverage~\citep{naeem2020reliable} for the marginal realism and diversity, respectively.}
    \label{suppl:metrics2}
\end{figure}

\begin{figure}[ht]
    \centering
    \includegraphics[width=\textwidth]{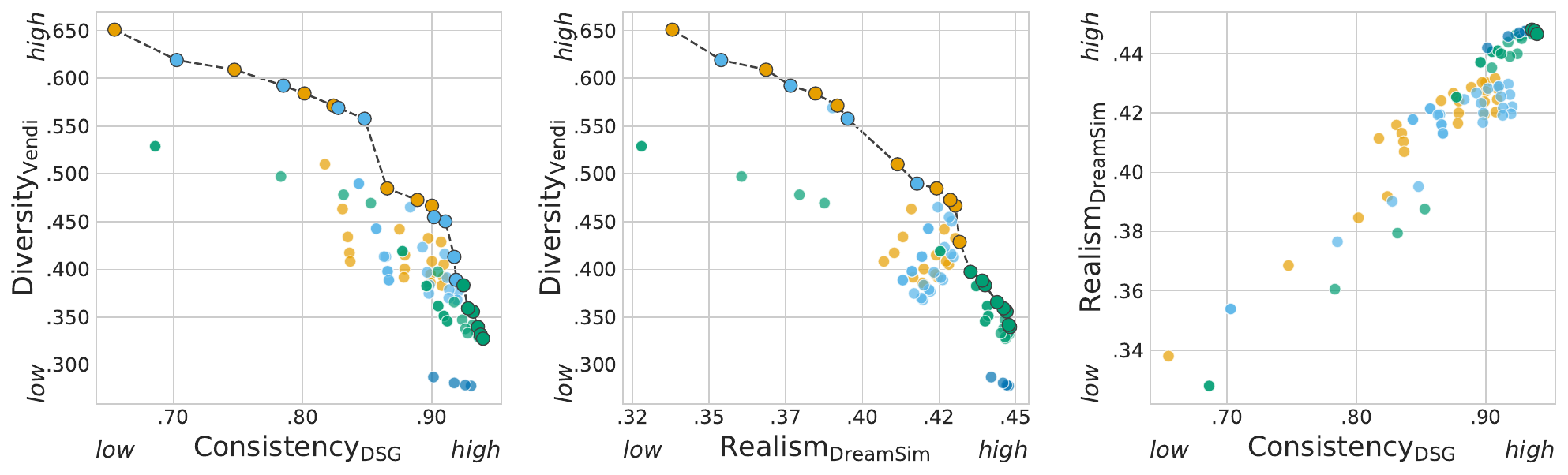}
    \includegraphics[width=\textwidth]{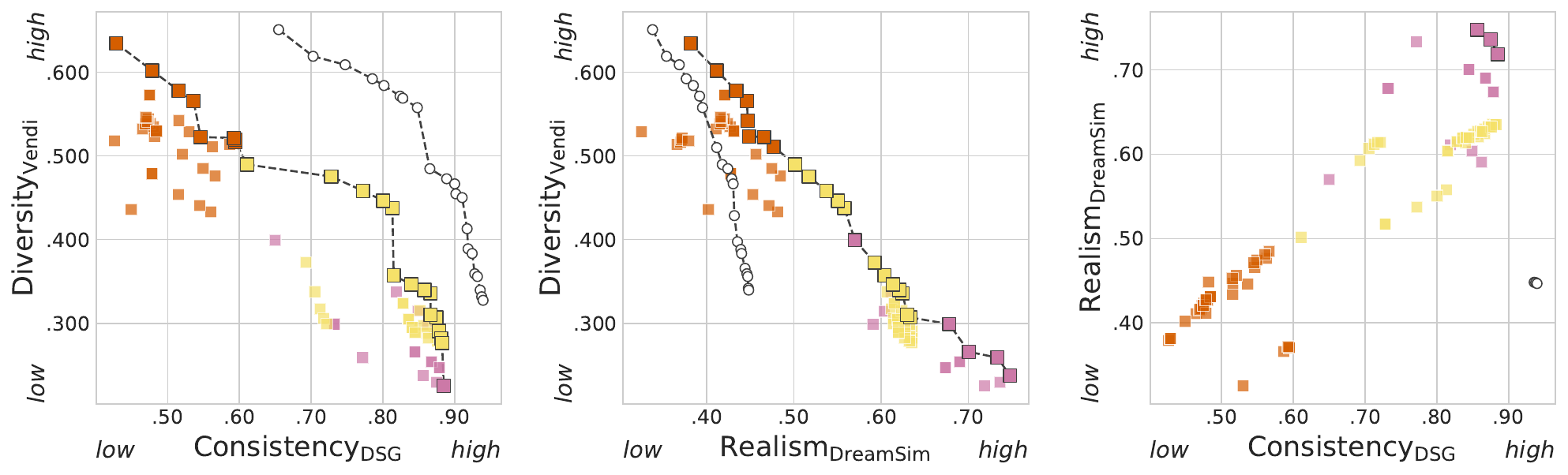}
    \caption{Using Vendi score~\citep{friedman2023vendi} for diversity.}
    \label{suppl:metrics4}
\end{figure}

\noindent\textbf{Additional metrics.} \cref{suppl:metrics1,suppl:metrics2,suppl:metrics4} ablate alternatives metrics for the consistency, diversity, and realism axes. We observe no major difference with respect to the Pareto fronts reported in the main paper.

\subsection{Additional results on GeoDE}
\begin{figure*}[ht]
    \centering
    
    \begin{subfigure}[b]{\textwidth}
    \includegraphics[width=.49\textwidth]{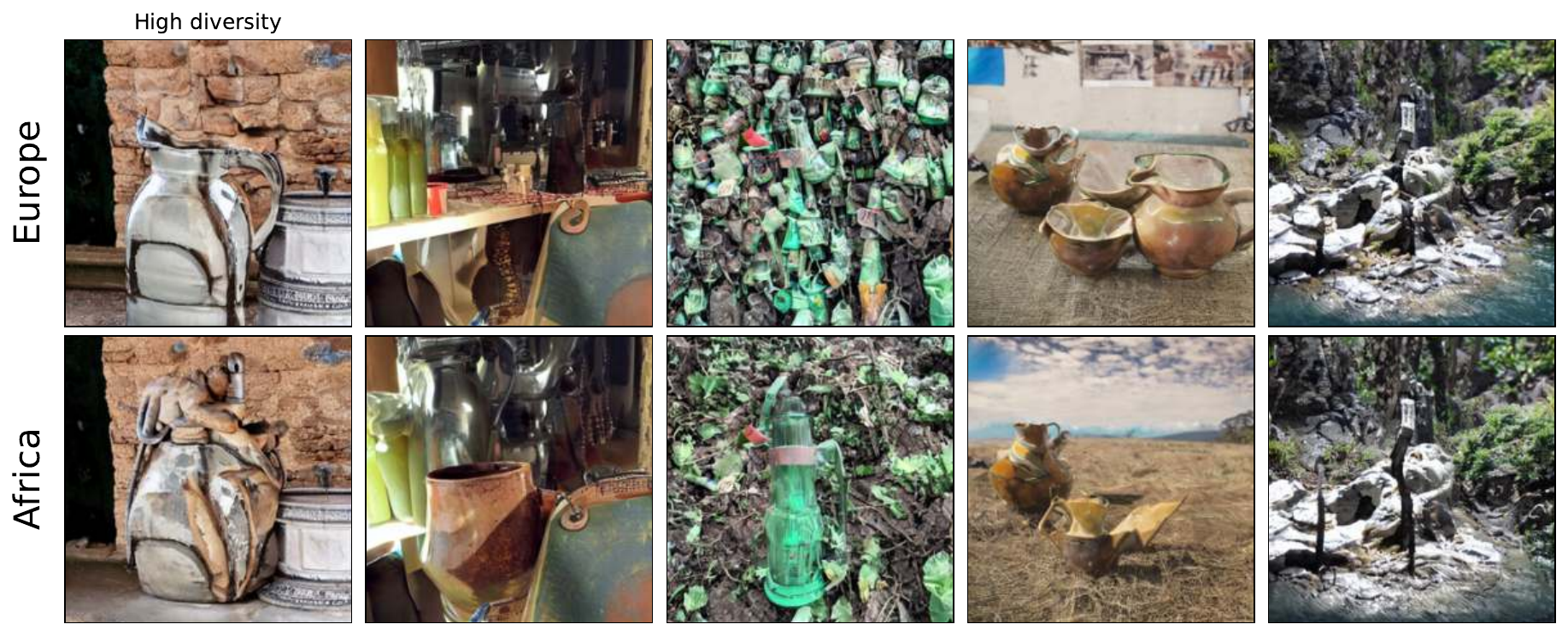}
    \includegraphics[width=.49\textwidth]{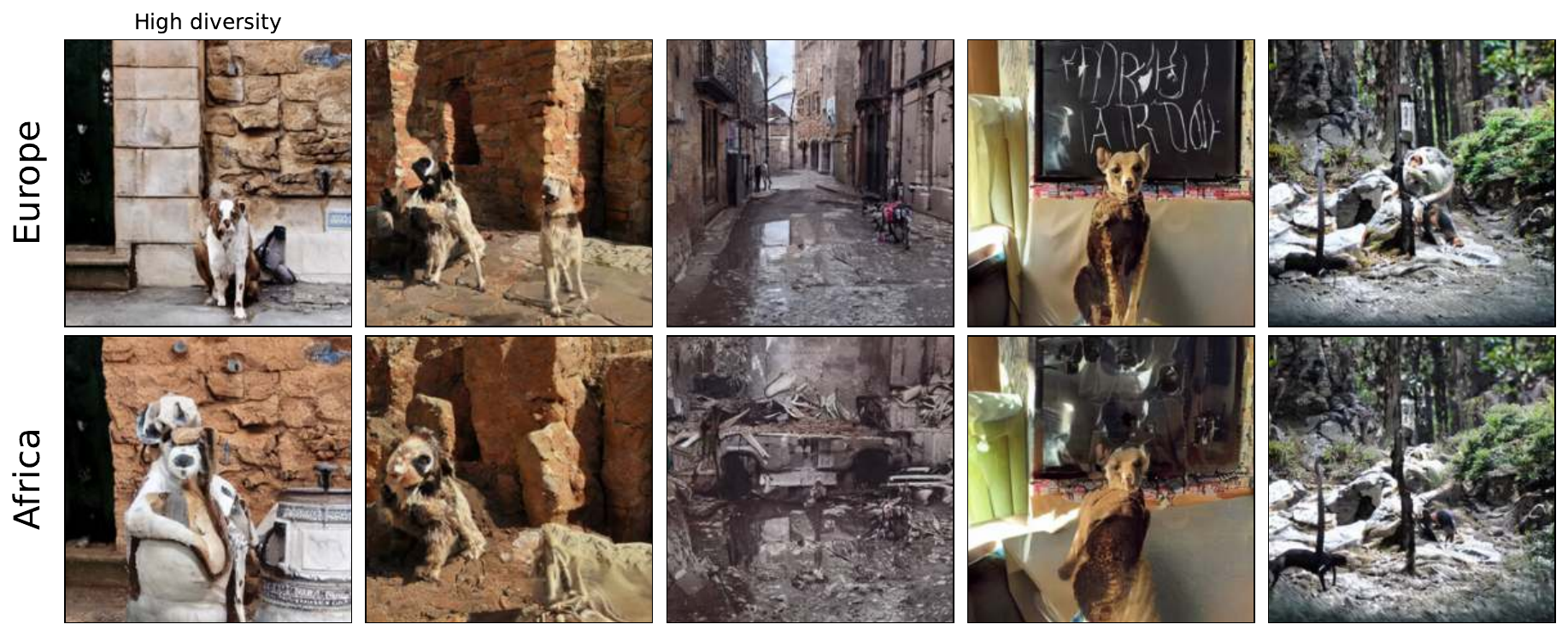}
    \caption{High diversity}
    \end{subfigure}

    \begin{subfigure}[b]{\textwidth}
    \includegraphics[width=.49\textwidth]{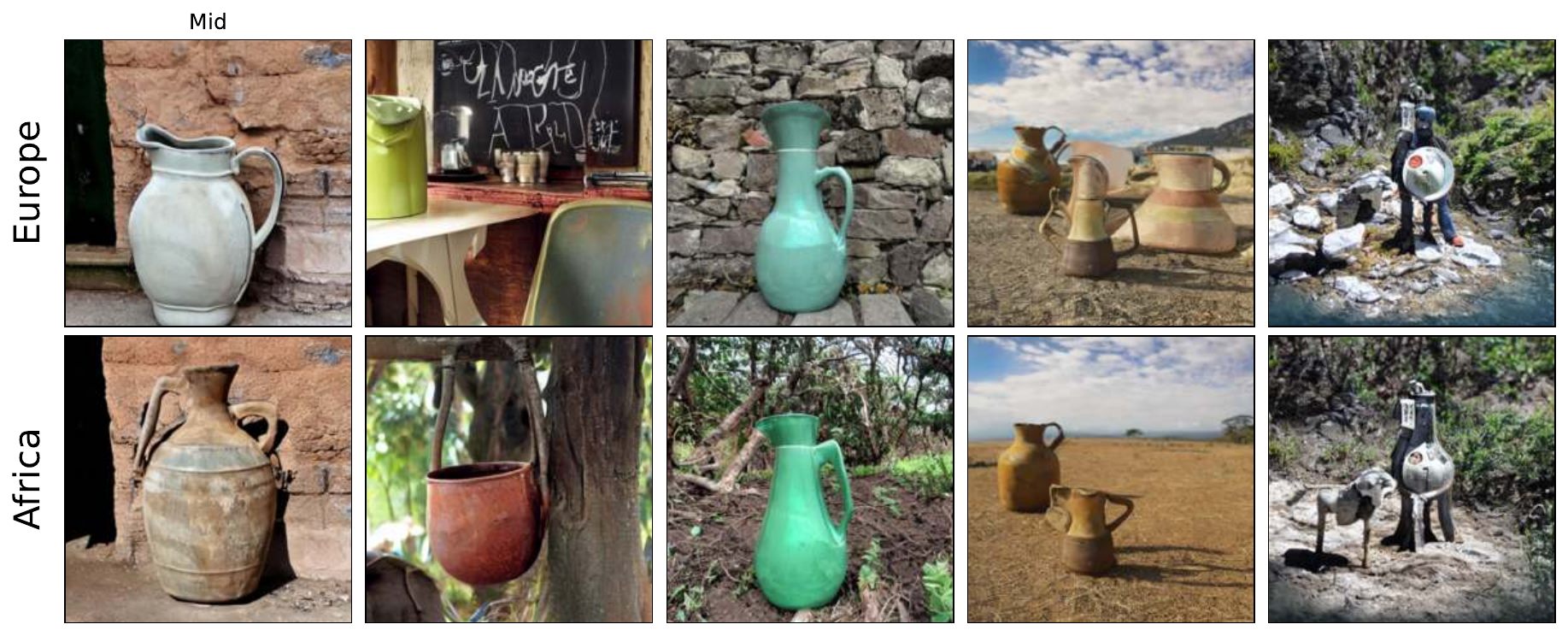}
    \includegraphics[width=.49\textwidth]{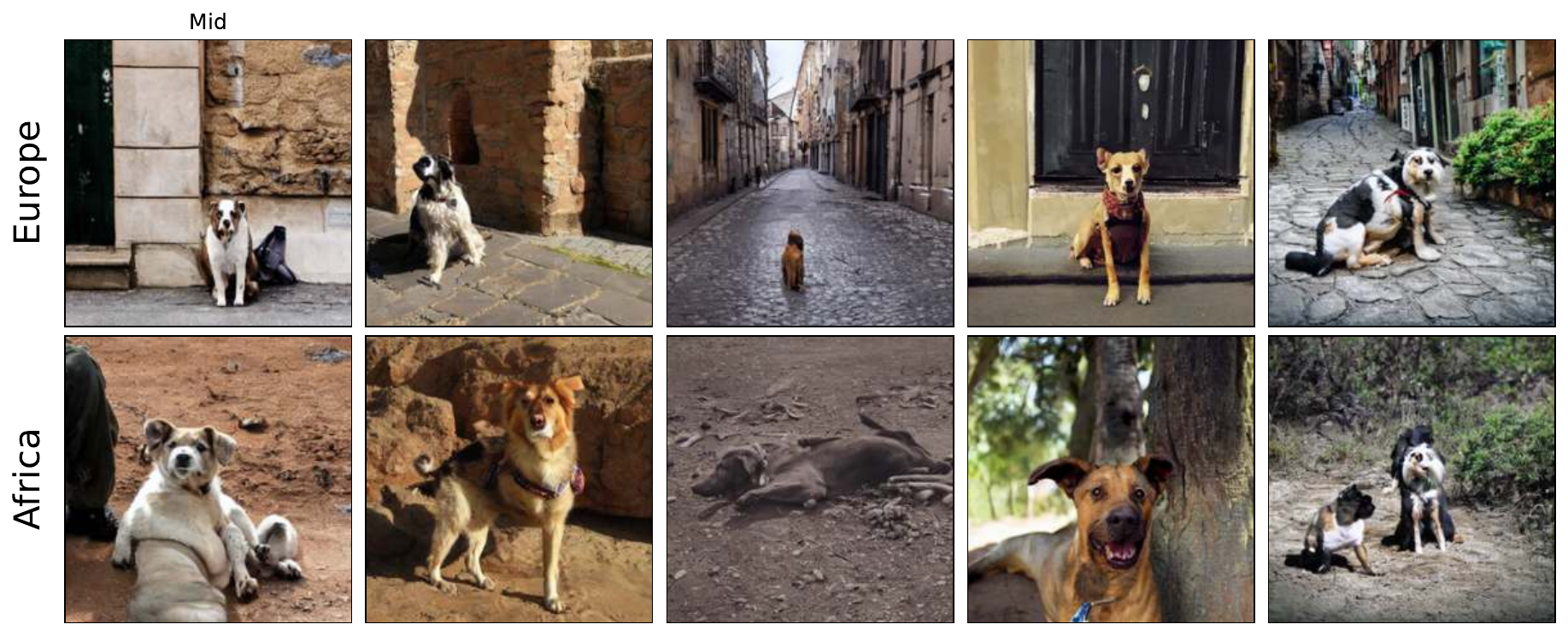}
    \caption{Balanced}
    \end{subfigure}
    
    \begin{subfigure}[b]{\textwidth}
    \includegraphics[width=.49\textwidth]{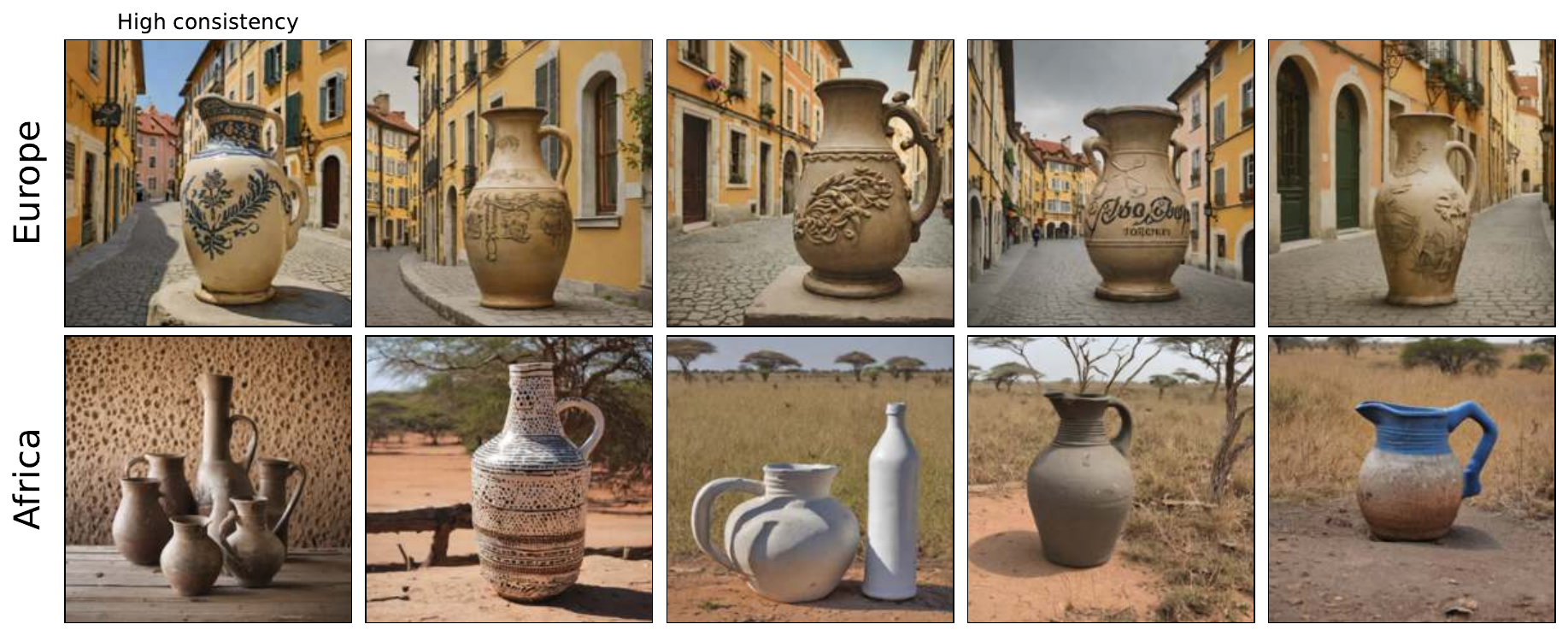}
    \includegraphics[width=.49\textwidth]{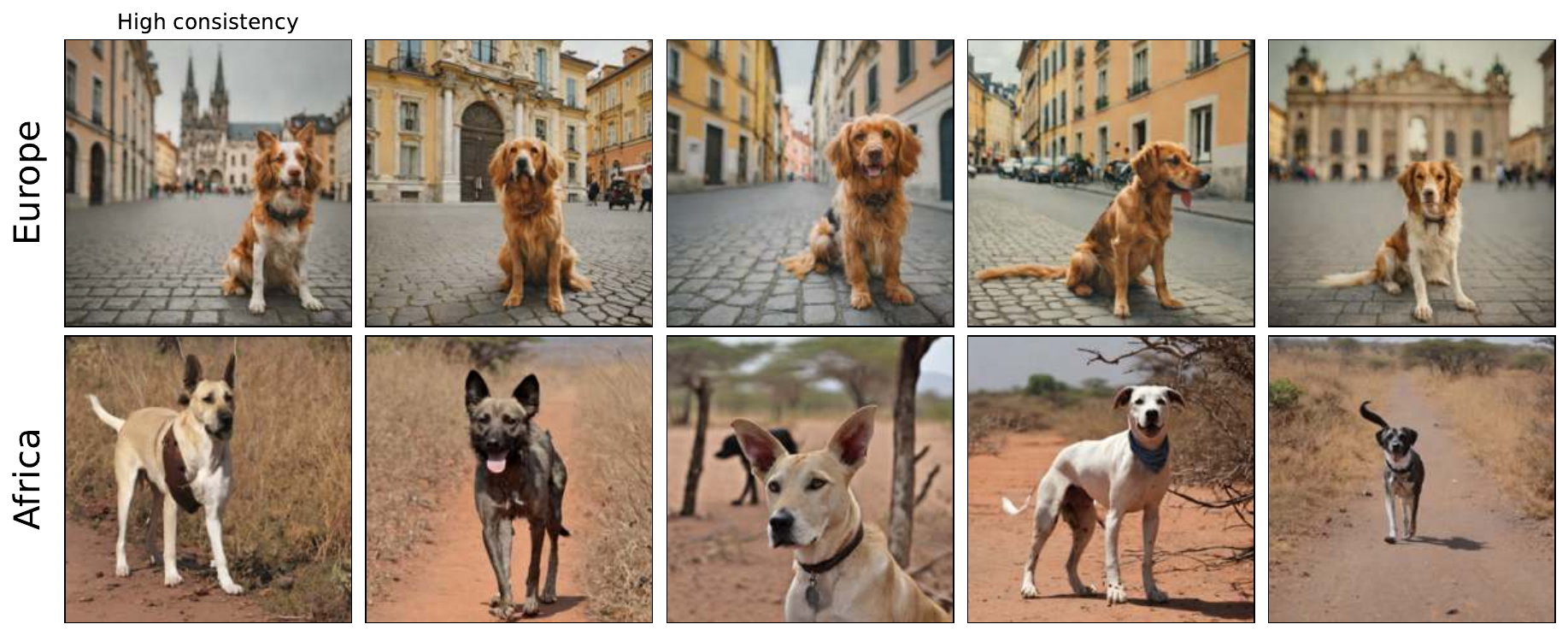}
    \caption{High consistency}
    \end{subfigure}

    \begin{subfigure}[b]{\textwidth}
    \includegraphics[width=.49\textwidth]{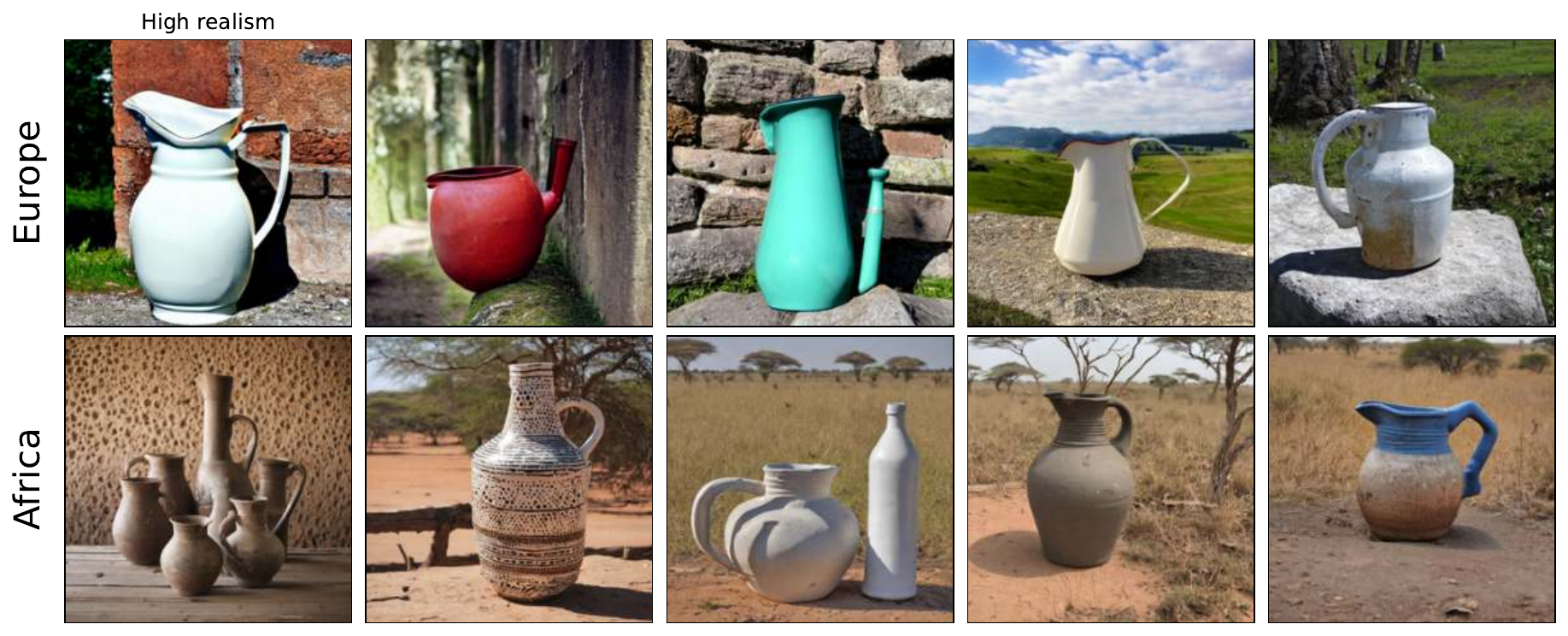}
    \includegraphics[width=.49\textwidth]{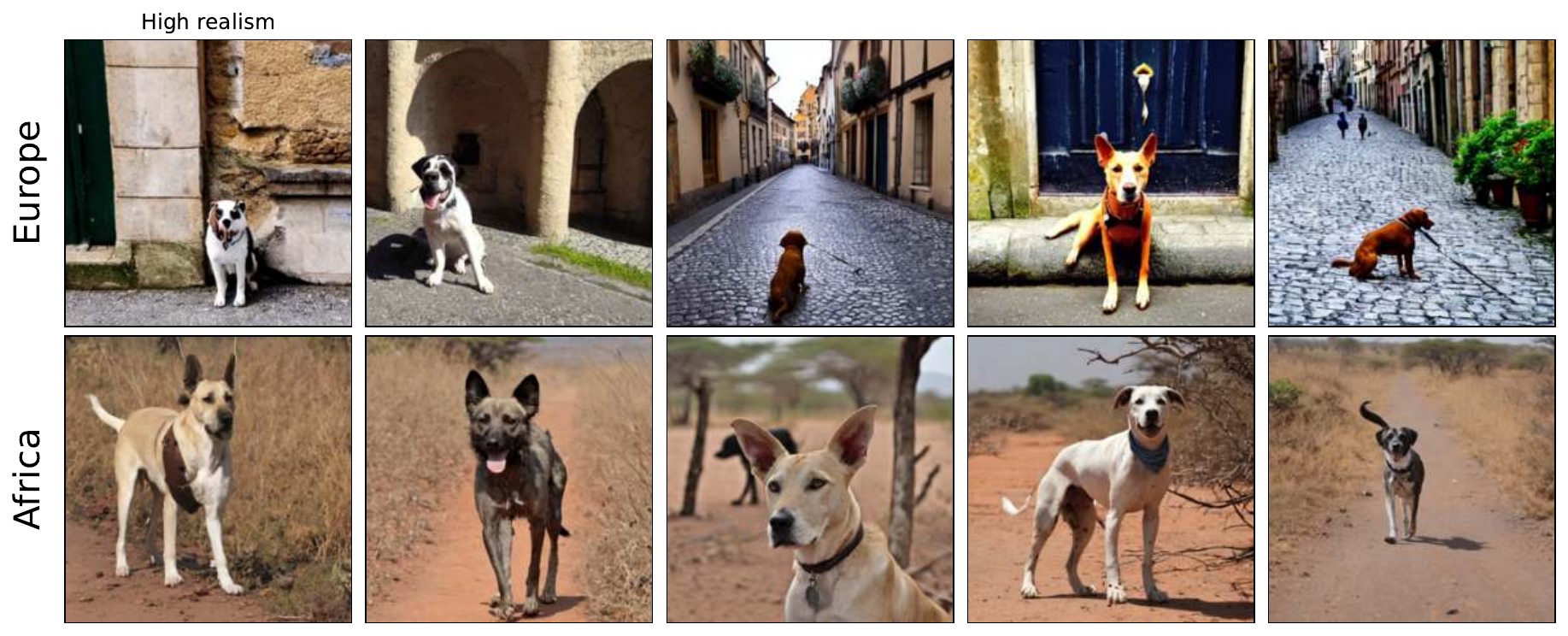}
    \caption{High realism}
    \end{subfigure}
    \caption{GeoDE qualitative.  Left: \texttt{A jug in \{region\}}. Right:  \texttt{A dog in \{region\}}}
    \label{suppl:quali_geode_pareto0}
\end{figure*}

\begin{figure*}[ht]
    \centering
    
    \begin{subfigure}[b]{\textwidth}
    \includegraphics[width=.49\textwidth]{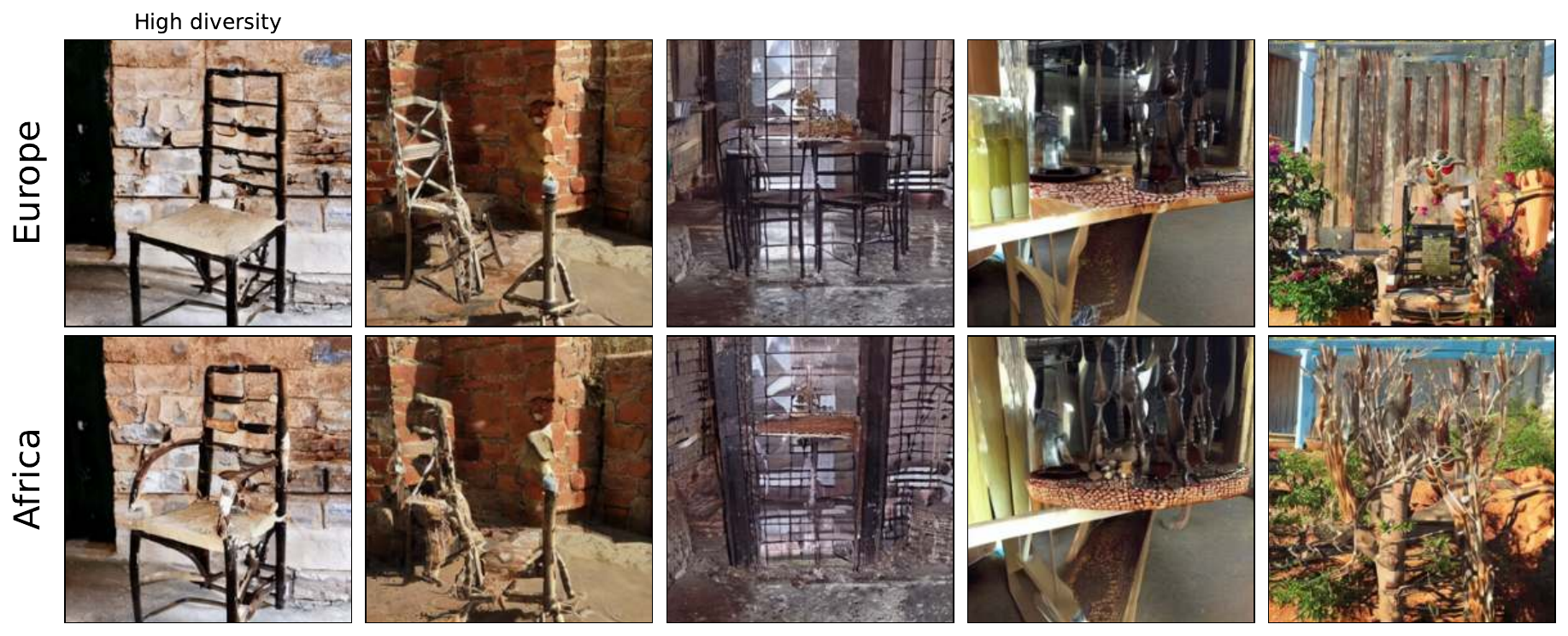}
    \includegraphics[width=.49\textwidth]{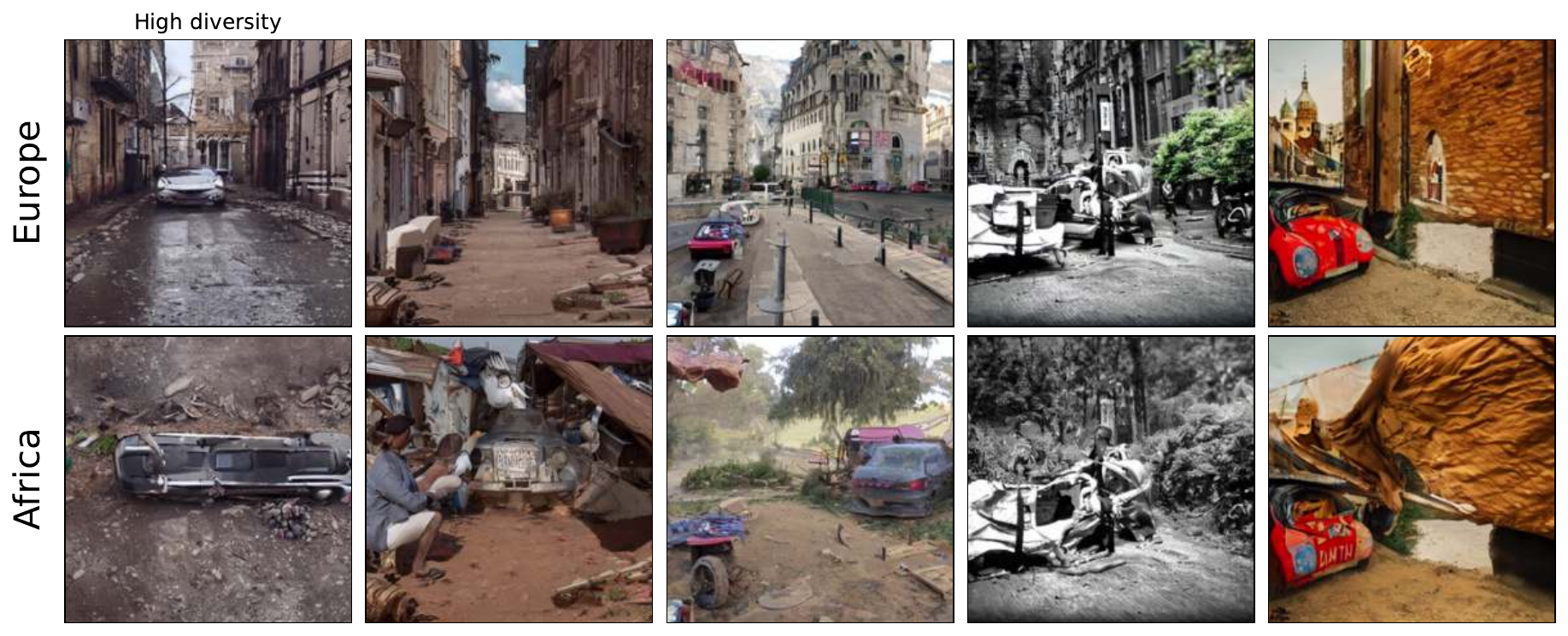}
    \caption{High diversity}
    \end{subfigure}

    \begin{subfigure}[b]{\textwidth}
    \includegraphics[width=.49\textwidth]{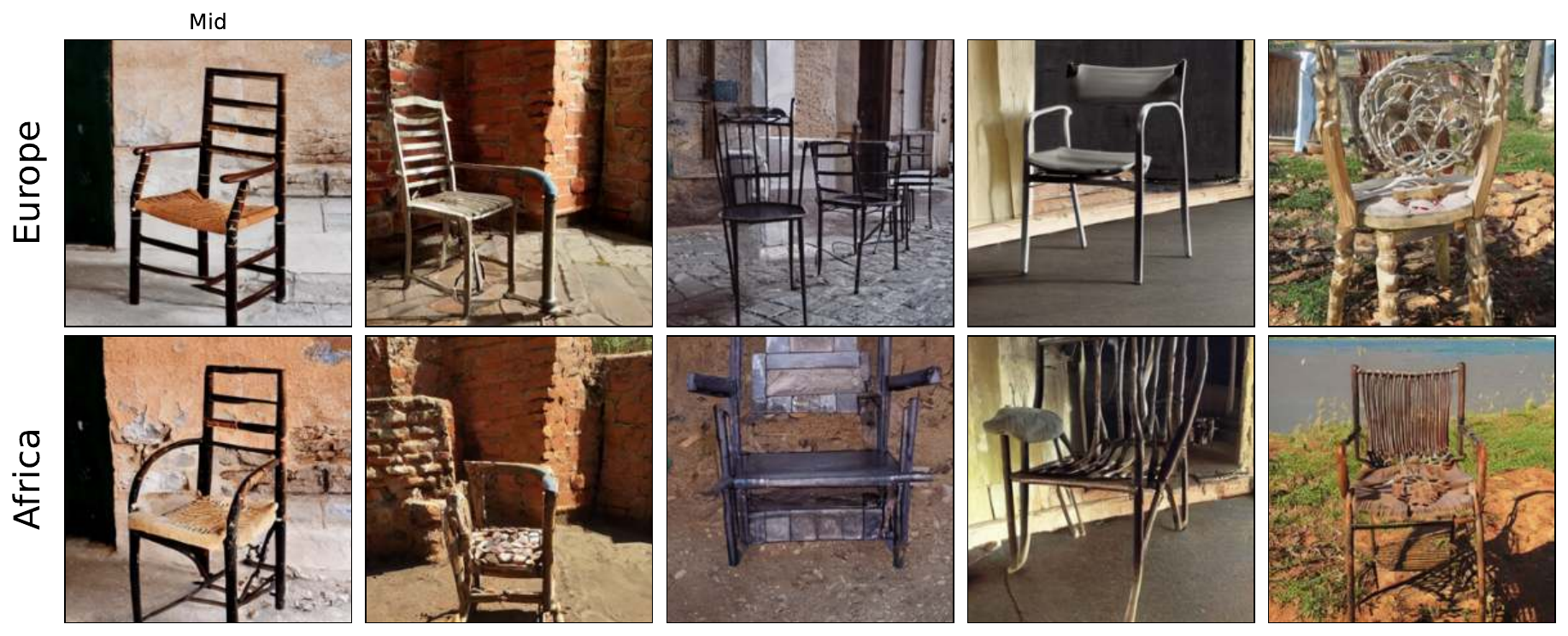}
    \includegraphics[width=.49\textwidth]{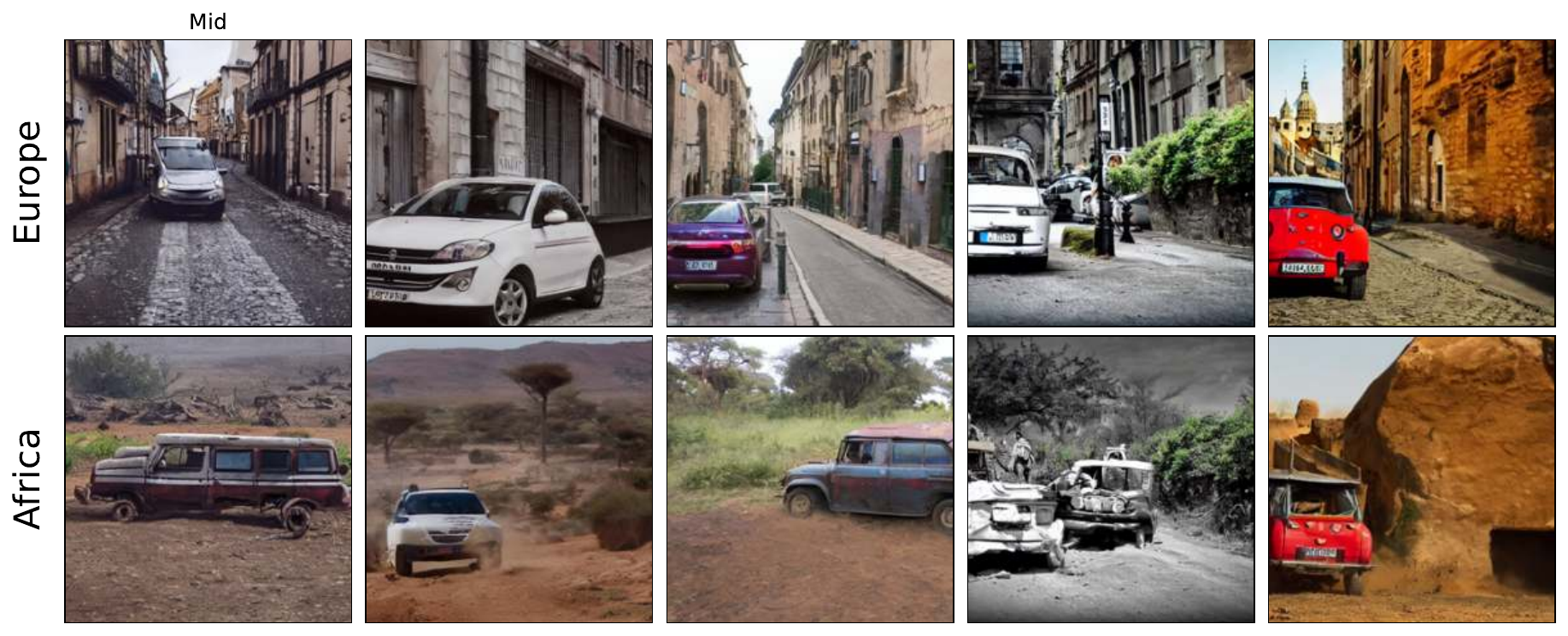}
    \caption{Balanced}
    \end{subfigure}
    
    \begin{subfigure}[b]{\textwidth}
    \includegraphics[width=.49\textwidth]{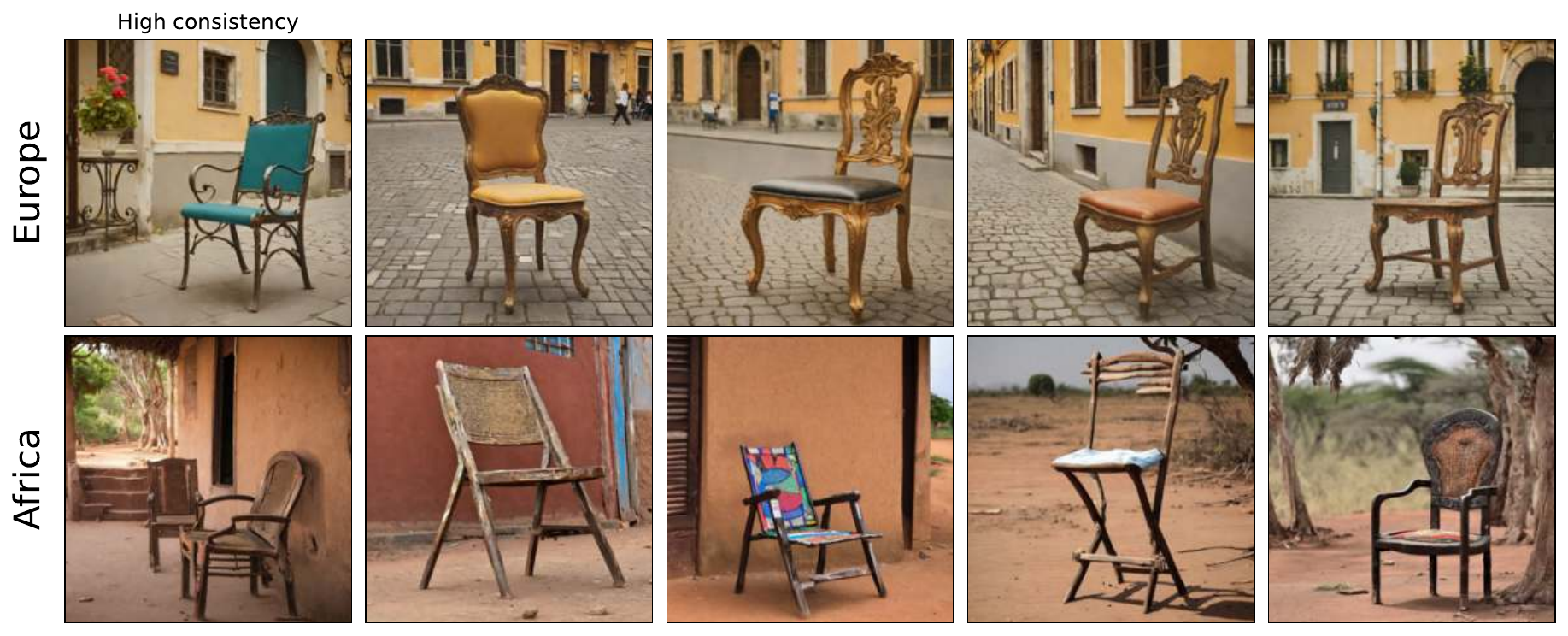}
    \includegraphics[width=.49\textwidth]{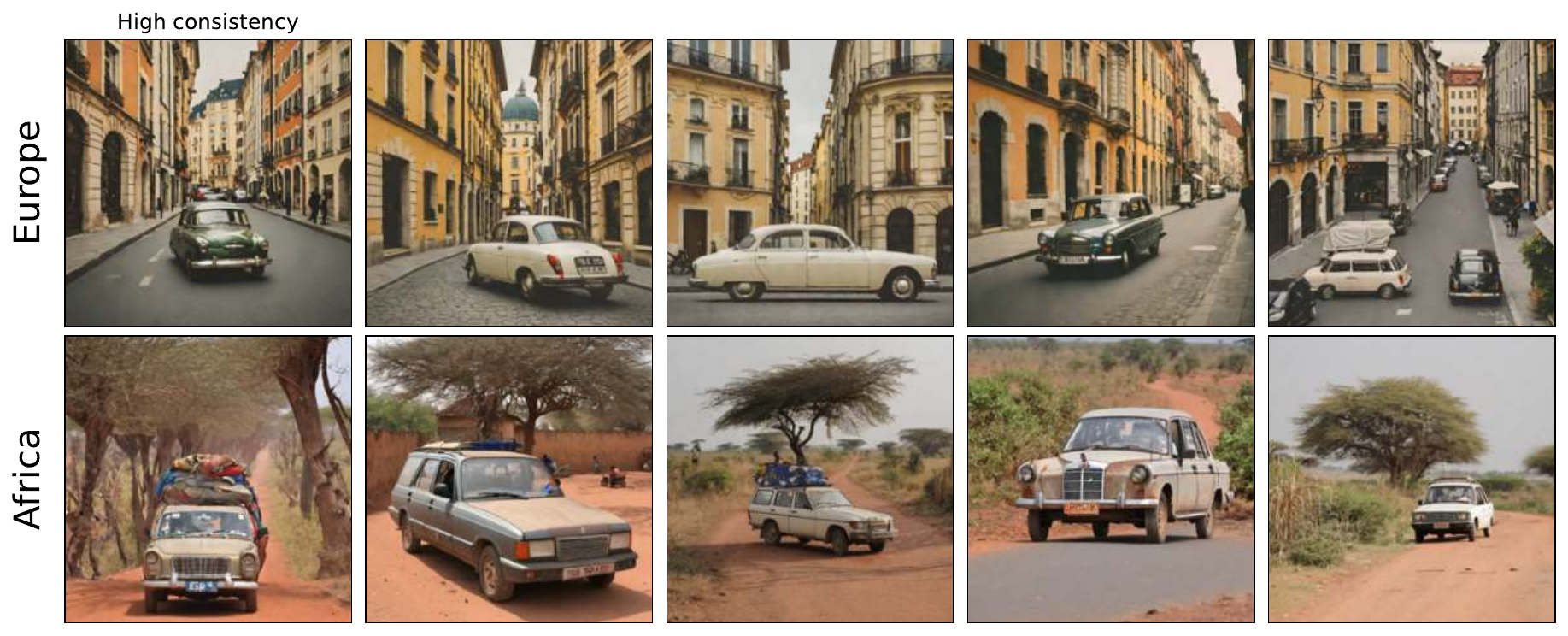}
    \caption{High consistency}
    \end{subfigure}

    \begin{subfigure}[b]{\textwidth}
    \includegraphics[width=.49\textwidth]{figures/qualitative/geode_pareto_id66_3.pdf}
    \includegraphics[width=.49\textwidth]{figures/qualitative/geode_pareto_id144_3.pdf}
    \caption{High realism.}
    \end{subfigure}
    \caption{GeoDE qualitative. Left: \texttt{A chair in \{region\}}. Right:  \texttt{A car in \{region\}}}
    \label{suppl:quali_geode_pareto1}
\end{figure*}

\noindent\textbf{Additional qualitative.} \cref{suppl:quali_geode_pareto0,suppl:quali_geode_pareto1} depict images generated with models present in the Pareto fronts at different locations. Four models are chosen in order to provide exemplars of different areas of the Pareto: one model has the highest diversity, one has balanced consistency-diversity or realism-diversity, one has the highest consistency, and one has the highest realism. By comparing the different block of rows, we notice that, as consistency and realism are increased, stereotypical generations of each region get exacerbated.

\end{document}